\documentclass[10pt,twocolumn,letterpaper]{article}

\usepackage[pagenumbers]{cvpr}

\usepackage{graphicx}
\usepackage{amsmath}
\usepackage{amssymb}
\usepackage{gensymb}
\usepackage{booktabs}

\usepackage{nicefrac}
\usepackage{multirow}
\usepackage{algpseudocode}
\usepackage{algorithm}
\usepackage{subcaption}
\usepackage{comment}
\usepackage[dvipsnames]{xcolor}

\usepackage{amsthm}
\newtheorem{theorem}{Theorem}

\usepackage[pagebackref,breaklinks,colorlinks]{hyperref}

\newtheorem{lemma}[theorem]{Lemma}

\newcommand{\x}{\mathbf{x}}

\newcommand{\MODELNAME}{BoDiffusion}

\usepackage[capitalize]{cleveref}
\crefname{section}{Sec.}{Secs.}
\Crefname{section}{Section}{Sections}
\Crefname{table}{Table}{Tables}
\crefname{table}{Tab.}{Tabs.}

\def\cvprPaperID{1780} 
\def\confName{CVPR}
\def\confYear{2023}

\newcommand\blfootnote[1]{%
\begingroup 
\renewcommand\thefootnote{}\footnote{#1}%
\addtocounter{footnote}{-1}%
\endgroup 
}

\begin{document}

\title{\MODELNAME: Diffusing Sparse Observations for Full-Body Human Motion Synthesis}

\author{
Angela Castillo$^{*1}$ \quad
Maria Escobar$^{*1}$ \quad
Guillaume Jeanneret$^{2}$ \quad
Albert Pumarola$^{3}$ \quad
Pablo Arbeláez$^{1}$ \\
Ali Thabet$^{3}$ \quad
Artsiom Sanakoyeu$^{3}$ \\
$^{1}$Center for Research and Formation in Artificial Intelligence, Universidad de los Andes \\ \quad $^{2}$University of Caen Normandie, ENSICAEN, CNRS, France \\
\quad $^{3}$Meta AI
}

\maketitle

\blfootnote{* Equal contributions.}
\begin{abstract}
    Mixed reality applications require tracking the user's full-body motion to enable an immersive experience. However, typical head-mounted devices can only track head and hand movements, leading to a limited reconstruction of full-body motion due to variability in lower body configurations. We propose \textbf{\MODELNAME}\ -- a generative diffusion model for motion synthesis to tackle this under-constrained reconstruction problem. We present a time and space conditioning scheme that allows \MODELNAME\ to leverage sparse tracking inputs while generating smooth and realistic full-body motion sequences. To the best of our knowledge, this is the first approach that uses the reverse diffusion process to model full-body tracking as a conditional sequence generation task. We conduct experiments on the large-scale motion-capture dataset AMASS and show that our approach outperforms the state-of-the-art approaches by a significant margin in terms of full-body motion realism and joint reconstruction error.

\end{abstract}

\section{Introduction}

\begin{figure}
  \centering
    \includegraphics[width=0.9\linewidth]{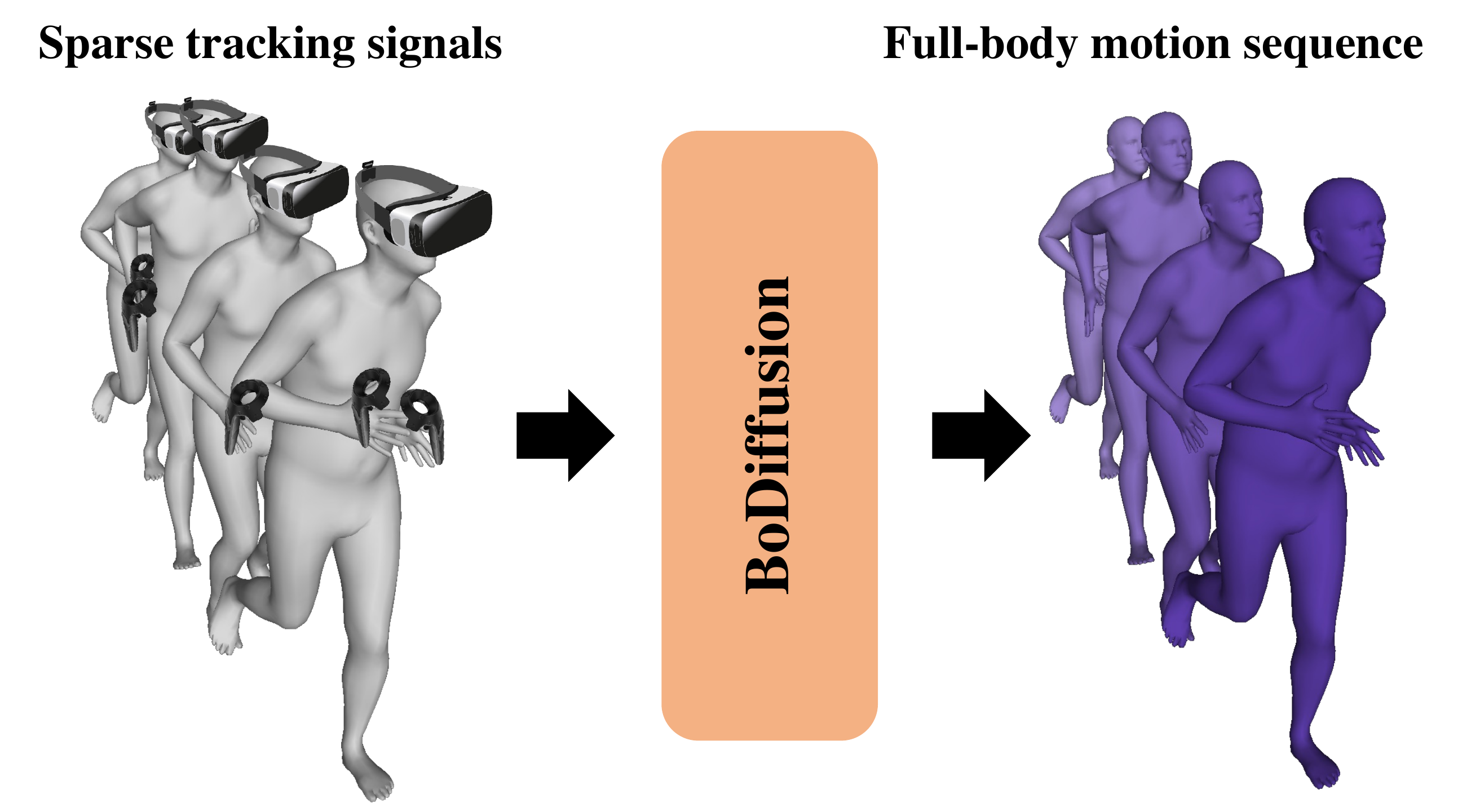}
  \caption{\textbf{\MODELNAME.} 
  Head and wrist IMUs are the standard motion-capture sensors in current virtual-reality devices. \MODELNAME~leverages the power of Transformer-based conditional Diffusion Models to synthesize fluid and accurate full-body motion from such sparse signals.
  }
  \label{fig:teaser}
\end{figure}

\begin{figure*}[t]
    \centering
    \includegraphics[width=\linewidth]{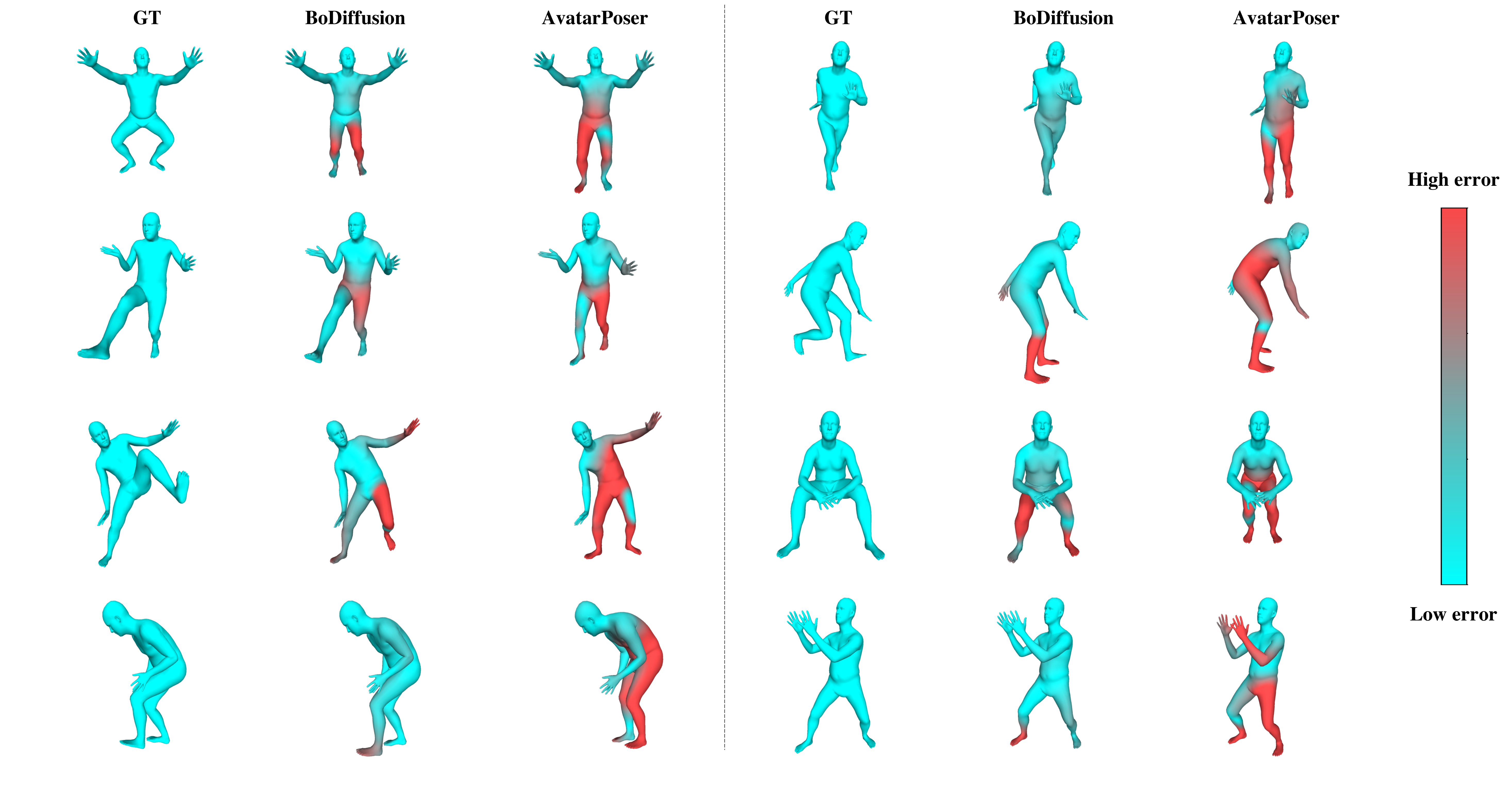}
    \caption{\textbf{Predicting Dense Full-Body Poses from Sparse Data.} 
    Comparison of \MODELNAME\ and AvatarPoser~\cite{jiang2022avatarposer} against the ground truth. Color gradient in the avatars indicates an absolute positional error, with a higher error corresponding to higher red intensity. \MODELNAME\ synthesizes substantially more accurate and plausible full-body poses, particularly in the lower body where no IMU data are captured.
    }
    \label{fig:qualitative_results_1}
\end{figure*}

Full-body motion capture enables natural interactions between real and virtual worlds for immersive mixed-reality experiences~\cite{jones2021belonging,piumsomboon2018mini,tecchia20123d}. 
Typical mixed-reality setups use a Head-Mounted Display (HMD) that captures visual streams with limited visibility of body parts and tracks the global location and orientation of the head and hands.
Adding more wearable sensors~\cite{huang2018deep,jiang2022transformer,kaufmann2021pose} is expensive and less comfortable to use.
Therefore, in this work, we tackle the challenge of enabling high-fidelity full-body motion tracking when only sparse tracking signals for the head and hands are available, as shown in Fig.~\ref{fig:teaser}. 

Existing motion reconstruction approaches for 3-point input (head and hands) struggle to model the large variety of possible lower-body motions and fail to produce smooth full-body movements because of their limited predictive nature~\cite{jiang2022avatarposer}. 
A recent attempt~\cite{aliakbarian2022flag} to address this problem uses a generative approach based on normalizing flows~\cite{rezende2015variational} falling short of incorporating temporal motion information and generating poses for every frame individually, thus resulting in unrealistic synthesized motions.
Another approach~\cite{dittadi2021full_VAE_HMD} that integrates motion history information using a Variational Autoencoder (VAE)~\cite{kingma2013auto} takes limited advantage of the temporal history because VAEs often suffer from ``posterior collapse"~\cite{he2018lagging,kingma2016improved}. 
Thus, there is a need for a scalable generative approach that can effectively model temporal dependencies between poses to address these limitations.

Recently, diffusion-based generative models~\cite{sohl2015deep,ho2020denoising} have emerged as a potent approach for generating data across various domains such as images~\cite{rombach2022high}, audio~\cite{Zhifeng2022DiffWave}, video~\cite{ho2022video}, and language~\cite{he2022diffusionbert}. Compared to Generative Adversarial Networks (GANs), diffusion-based models have demonstrated to capture a much broader range of the target distribution~\cite{nichol2021improved}. They offer several advantages, including excellent log-likelihoods and high-quality samples, and employ a solid, stationary training objective that scales effortlessly with training compute~\cite{nichol2021improved}.

To leverage the powerful diffusion model framework, we propose \textbf{\MODELNAME} (\textbf{Bo}dy \textbf{Diffusion}), a new generative model for human motion synthesis. \MODELNAME\ directly learns the conditional data distribution of human motions, models temporal dependencies between poses, and generates full \emph{motion sequences}, in contrast to previous methods that operate solely on static poses~\cite{aliakbarian2022flag,yang2021lobstr}. 
Moreover, \MODELNAME\ does not suffer from the limitation of methods that require a known pelvis location and rotation during inference~\cite{dittadi2021full_VAE_HMD,aliakbarian2022flag,yang2021lobstr}, and generates high-fidelity body motions relying solely on the head and hands tracking information. 

Our main contributions can be summarized as follows.
We propose \MODELNAME\ -- the first diffusion-based generative model for full-body motion synthesis conditioned on the sparse tracking inputs obtained from HMDs.
To build our diffusion model, we adopt a Transformer-based backbone~\cite{Peebles2022DiT}, which has proven more efficient for image synthesis than the frequently used UNet backbone \cite{Dhariwal2021DiffusionMB,ramesh2022hierarchical,rombach2022high}, and it is more naturally suited for modeling sequential motion data.
To enable conditional motion synthesis in \MODELNAME, we introduce a novel time and space conditioning scheme, where global positions and rotations of tracked joints encode the control signal. 
Our extensive experiments on AMASS~\cite{AMASS:ICCV:2019} demonstrate that the proposed \MODELNAME\ synthesizes smoother and more realistic full-body pose sequences from sparse signals, outperforming the previous state-of-the-art methods (see Fig.~\ref{fig:qualitative_results_1} and~\ref{fig:qualitative_results_2}). 
Find our full project on \href{https://bcv-uniandes.github.io/bodiffusion-wp/}{bcv-uniandes.github.io/bodiffusion-wp/}.


\section{Related Work}
\begin{figure*}[ht!]
    \centering
    \includegraphics[width=0.9\linewidth]{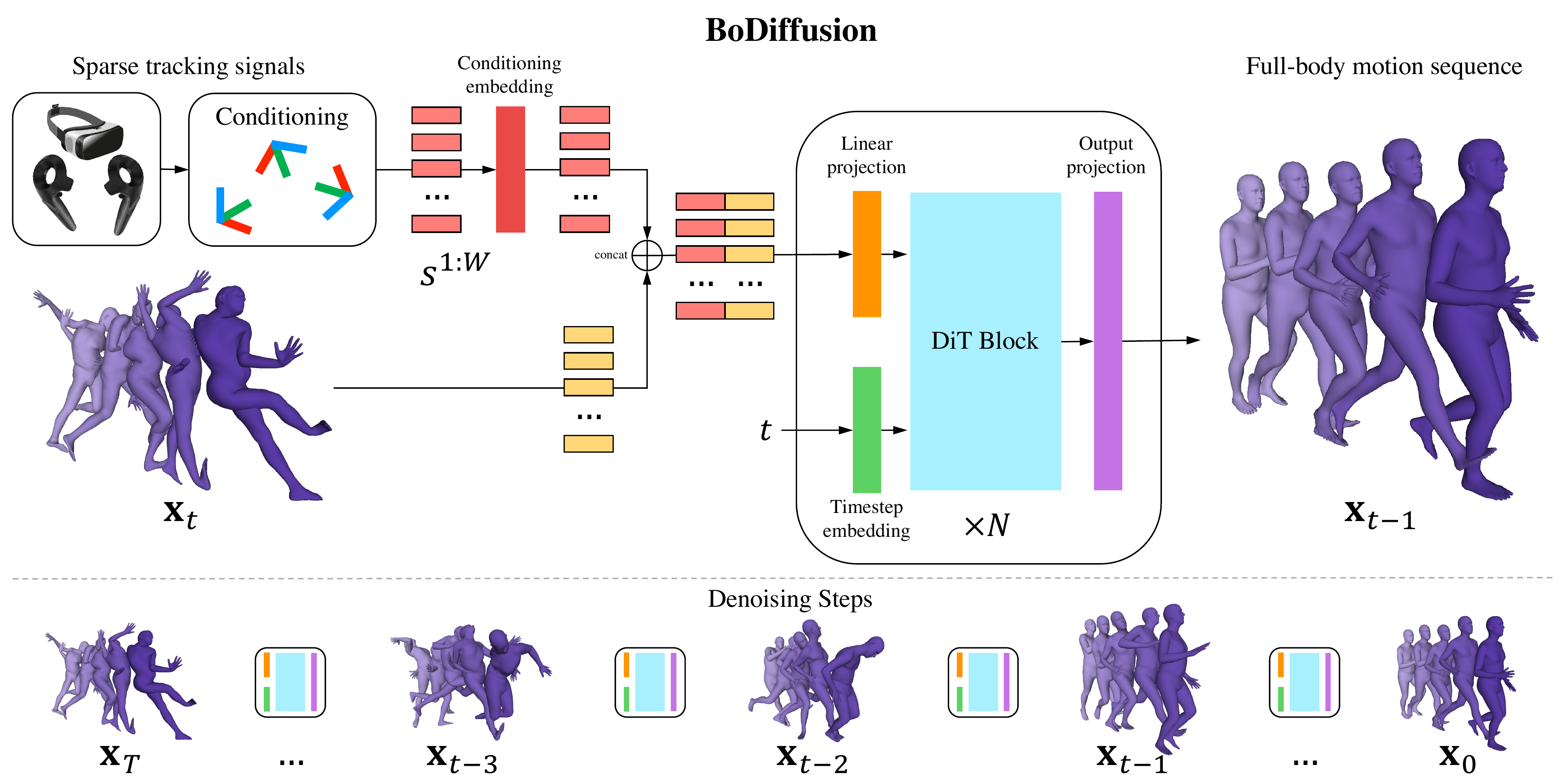}
    \caption{\textbf{Overview of \MODELNAME.} 
    \MODELNAME\ is a diffusion process synthesizing full-body motion using sparse tracking signals as conditioning. \textbf{Top:} At each denoising step, the model takes as input $2W$ tokens, which correspond to local joint rotations with $t$ steps of noise ($\x_t=x_t^{1:W}$) and sparse tracking signals of the head and hands ($s^{1:W}$) as conditioning. We concatenate the $x_t^i$ tokens with the conditioning tokens $s^i$ along the spatial axis to preserve the time information and ensure coherence between the conditioning signal and the synthesized motion. After that, we pass it through the Transformer backbone of $N$ DiT blocks~\cite{Peebles2022DiT}. \textbf{Bottom:} During inference, we start from random Gaussian noise $\x_T$ and perform $T$ denoising steps until we reach a clean output motion $\x_0$.
    }
    \label{fig:pipeline}
\end{figure*}

\paragraph{Pose Estimation from Sparse Observations.} Full-body pose estimation methods generally rely on inputs from body-attached sensors. Much prior work relies on 6 Inertial Measurement Units (IMUs) to predict a complete pose~\cite{huang2018deep,PIPCVPR2022,TransPoseSIGGRAPH2021}. In~\cite{huang2018deep}, the authors train a bi-directional LSTM to predict body joints of a SMPL~\cite{SMPL:2015} model, given 6 IMU inputs (head, 2 arms, pelvis, and 2 legs). However, there is a high incentive to reduce the number of body-attached IMUs because depending on many body inputs creates friction in motion capture. LoBSTr~\cite{yang2021lobstr} reduces this gap by working with 4 inputs (head, 2 arms, and pelvis). It takes past tracking signals of these body joints as input for a GRU network that predicts lower-body pose at the current frame. Furthermore, it estimates the upper body with an Inverse Kinematics (IK) solver. The methods in~\cite{aliakbarian2022flag,dittadi2021full_VAE_HMD} also require 4 joints as input since they leverage the pose of the pelvis to normalize the input data during training and inference.

In Mixed Reality (MR), obtaining user input from a headset and a pair of controllers is common.
The authors of~\cite{jiang2022avatarposer,winkler2022questsim} highlight the importance of a sensor-light approach and further reduce the amount of inputs to 3, a number that aligns well with scenarios in MR environments. AvatarPoser~\cite{jiang2022avatarposer} combines a Transformer architecture and traditional IK to estimate full-body pose from HMD and controller poses. Similar to~\cite{jiang2022avatarposer}, our method uses only 3 inputs but provides much better lower-body prediction thanks to our diffusion model. Choutas~\etal~\cite{choutas2022learning} propose an iterative neural optimizer for 3D body fitting from sparse HMD signals. However, they optimize poses frame-by-frame and do not consider motions. QuestSim~\cite{winkler2022questsim} proposes to learn a policy network to predict joint torques and reconstruct full body pose using a physics simulator. Nevertheless, this approach is challenging to apply in a real-world scenario, especially when motion involves interaction with objects (\emph{e.g.}, sitting on a chair). 
In such a case, one needs to simulate both the human body and all the objects, which have to be pre-scanned in advance and added to the simulation.
In contrast, our approach is data-driven and does not require a costly physics simulation or object scanning. 


\paragraph{Human Motion Synthesis \& Pose Priors.} A large body of work aims at generating accurate human motion given no past information~\cite{ahn2018text2action,petrovich2021action,zanfir2020weakly,petrovich2022temos,lin2022ohmg}. Methods like TEMOS~\cite{petrovich2022temos} and OhMG~\cite{lin2022ohmg} combine a VAE~\cite{kingma2013auto} and a Transformer network to generate human motion given text prompts. Recently, FLAG~\cite{aliakbarian2022flag} argues against the reliability of using VAEs for body estimation and proposes to solve these disadvantages with a flow-based generative model. VPoser~\cite{pavlakos2019expressive} learns a pose prior using VAE, and Humor~\cite{rempe2021humor} further improves it by learning a conditional prior using a previous pose. Recent work~\cite{oreshkin2021protores} proposes a more generic approach that learns a pose prior and approximates an IK solver using a neural network.
Another line of work tackles motion synthesis using control signals provided by an artist or from game-pad input~\cite{holden2017phase,ling2020character,henter2020moglow,peng2021amp,starke2022deepphase}. However, in contrast to our method, such approaches either focus on locomotion and rely on the known future root trajectory of the character or are limited to a predefined set of actions~\cite{peng2021amp}.

\paragraph{Denoising Diffusion Probabilistic Models (DDPMs)~\cite{ho2020denoising,nichol2021improved}} are a class of likelihood-based generative models inspired by Langevin dynamics~\cite{langevin1908theorie} which map between a prior distribution and a target distribution using a gradual denoising process. Specifically, generation starts from a noise tensor and is iteratively denoised for a fixed number of steps until a clean data sample is reached. Recently, Ho~\etal~\cite{ho2020denoising} have shown~\cite{ho2020denoising} that DDPMs are equivalent to the score-based generative models~\cite{song2019generative,song2020improved}. Currently, DDPMs are showing impressive results in tasks like image generation and manipulation~\cite{Dhariwal2021DiffusionMB,rombach2022high,ramesh2022hierarchical,gafni2022make,nichol2021glide} due to their impressive ability to fit the training distribution at large scale and stable training objective. 
Moreover, concurrent to this work, Diffusion Models have also been used to synthesize human motion from text inputs~\cite{zhang2022motiondiffuse,kim2022flame,tevet2022human}. 

UNet~\cite{ronneberger2015u} architecture has been de-facto the main backbone for image synthesis with Diffusion Models~\cite{Dhariwal2021DiffusionMB,ramesh2022hierarchical,rombach2022high} up until a recent work~\cite{Peebles2022DiT} that suggested a new class of DDPMs for image synthesis with Transformer-based backbones. Transformers are inherently more suitable than convolutional networks for modeling heterogeneous sequential data, such as motion, and we capitalize on this advantage in our work. In particular, we employ a Transformer-based Diffusion Model, based on the DiT backbone~\cite{Peebles2022DiT}, to construct an architecture for conditional full-body pose estimation from 3 IMU tracking inputs.

\section{\MODELNAME}
In this section, we present our \MODELNAME\ model. We start with the DDPMs background in Sect.~\ref{sec:diffusion_background}. Next, we define the problem statement and our probabilistic framework in Sect.~\ref{sec:cond_full_body_motion_synthesis}. Then, in Sect.~\ref{sec:bodiffusion_arch}, we give an overview of the proposed \MODELNAME\ model for conditional full-body motion synthesis from sparse tracking signals, followed by the details of our model design. Please refer to Fig.~\ref{fig:pipeline} for an illustration of the entire pipeline of our method. 

\subsection{Diffusion Process}\label{sec:diffusion_background}

We briefly summarize DDPMs~\cite{ho2020denoising} inner workings and formulate our conditional full-body motion synthesis task using the generative framework.
Let $x_0^{1:W}=\x_0 \sim q(\x_0)$ be our real motion data distribution, where $W$ is the length of the sequence motion. The forward diffusion process $q$ produces latent representations $\x_1,\dots, \x_T$ by adding Gaussian noise at each timestep $t$ with variances $\beta_t \in (0, 1)$. Hence, the data distribution is defined as follows:
\begin{equation}
    q(\x_{1:T} | \x_0) = \prod_{t=1}^T q(\x_t | \x_{t-1})
\end{equation}
\begin{equation}
    q(\x_{t}| \x_{t-1})=\mathcal{N}(\x_t; \sqrt{1 - \beta_t} \x_{t-1}, \beta_t\mathbf{I}),
\end{equation}
where $\mathbf{I}$ is the identity matrix. 
Due to the properties of Gaussian distributions, Ho~\etal~\cite{ho2020denoising} showed that we can directly calculate $\x_t$ from $\x_0$ by sampling:
\begin{equation}\label{eq:x_t}
    \x_t = \sqrt{\bar{\alpha}_t} \x_0 + \sqrt{1 - \bar{\alpha}_t} \epsilon,
\end{equation}
where $\alpha_t = 1 - \beta_t$, $\bar{\alpha_t} = \prod_{i=1}^T\alpha_i$, and $\epsilon \sim \mathcal{N}(0, \mathbf{I})$.

On the contrary, the reverse diffusion process $q(\x_{t-1}| \x_{t})$ is the process of iterative denoising through steps $t=T,\dots, 1$. Ideally, we would like to perform this process in order to convert Gaussian noise $\x_T \sim \mathcal{N}(0, \mathbf{I})$ back to the data distribution and generate real data points $\x_0$. However, $q(\x_{t-1}| \x_{t})$ is intractable because it needs to use the entire data distribution. Therefore, we approximate it with a neural network $p_\theta$ with parameters $\theta$:
\begin{equation}\label{eq:p_inverse}
    p_{\theta}(\x_{t-1}| \x_t) = \mathcal{N}(\x_{t-1}; \mu_{\theta} (\x_t, t), \Sigma_\theta(\x_t, t)).  
\end{equation}
We train to optimize the negative log-likelihood using the Variational Lower Bound (VLB) ~\cite{ho2020denoising}:
\begin{equation}\label{eq:loss_vlb}
\begin{split}
- \log\ &p_\theta(\x_0) 
\leq - \log p_\theta(\x_0) + \\
&+D_\text{KL}(q(\x_{1:T}\vert \x_0) \| p_\theta(\x_{1:T}\vert \x_0) ) = \mathcal{L}_{\text{vlb}}.
\end{split}
\end{equation}

Following~\cite{ho2020denoising}, we parameterize $\mu_{\theta} (\x_t, t)$ like this:
\begin{equation}\label{eq:objective-mean}
    \mu_{\theta} (\x_t, t) = \frac{1}{\sqrt{\alpha_t}}\left(\x_t - \frac{\beta_t}{\sqrt{1 - \bar{\alpha}_t}}\epsilon_{\theta}(\x_t, t) \right).
\end{equation}

After a couple simplifications, \cite{ho2020denoising} ignores the weighting terms to rewrite $\mathcal{L}_{\text{simple}}$ as follows:

\begin{equation}\label{eq:lsimple}
    \mathcal{L}_{\text{simple}} = E_{\x_0 \sim q(\x_0), t \sim U[1,T]}||\epsilon - \epsilon_{\theta}(\x_t, t) ||^2_2.
\end{equation}
Ho~\etal~\cite{ho2020denoising} observed that optimizing $\mathcal{L}_{\text{simple}}$ works better in practice than optimizing full VLB $\mathcal{L}_{\text{vlb}}$.
During training, we follow Eq.~\ref{eq:lsimple}, where we sample $\x_0$ from the data distribution, the timestep as $t \sim 
\mathcal{U}\{1,T\}$, and compute $\x_t$ using Eq.~\ref{eq:x_t}. 
Intuitively, we learn $p_{\theta}(\x_{t-1}| \x_t)$ by training neural network to predict the noise $\epsilon$ that was used to compute the $\x_t$ with Eq.~\ref{eq:x_t}. 
However, simple loss $\mathcal{L}_{\text{simple}}$ assumes that we have a predefined variance $\Sigma(\x_t, t)=\beta_t$. Instead, we follow \cite{nichol2021improved} and optimize the variance 
$\Sigma_\theta(\x_t, t) = e^{v \log \beta_t + (1-v) \log \beta_t \frac{1 - \bar{\alpha}_{t-1}}{1 - \bar{\alpha}_1}}$,
where $v$ is a learnable scalar. Hereby, we use a combined objective:
\begin{equation}
    \mathcal{L} =\mathcal{L}_{\text{simple}} + \lambda_{\text{vlb}} \mathcal{L}_{\text{vlb}}.
\end{equation}

\subsection{Conditional Full-Body Motion Synthesis}\label{sec:cond_full_body_motion_synthesis}
\textbf{Problem Definition.} Human motion can be characterized by a sequence of body poses $x^i$ ordered in time. We define a~\emph{pose} as a set of body joints arranged in the kinematic tree of the SMPL~\cite{SMPL:2015} model. 
Joint states are described by their local rotations relative to their parent joints, with the pelvis serving as the root joint and its rotation being defined with the global coordinate frame. We utilize the 6D representation of rotations~\cite{zhou2019continuity} to ensure favorable continuity properties, making $x^i \in \mathbb{R}^{22 \times 6}$. 
The global translation of the pelvis is not modeled explicitly, as it can be calculated from the tracked head position by following the kinematic chain~\cite{jiang2022avatarposer}.
We consider a typical mixed reality system with HMD and two hand controllers that provides~\emph{3-point} tracking information of head and hands in the form of their global positions $p^i$ and rotations $r^i$.
Furthermore, we additionally compute the linear and angular velocities $v^i, \omega^i$ of the head and wrists, making $s^i = \{p^i, r^i, v^i, \omega^i\} \in \mathbb{R}^{3 \times (3+6+3+6)}$ to make the input signal more informative and robust~\cite{jiang2022avatarposer}. 
The target task is to synthesize full-body human motion ${x^{1:W}=\{x^i\}_{i=1}^W}$ using the limited tracking signals ${s^{1:W}=\{s^i\}_{i=1}^W}$ as input.

\begin{figure}[t]
    \centering
    \includegraphics[width=0.8\linewidth]{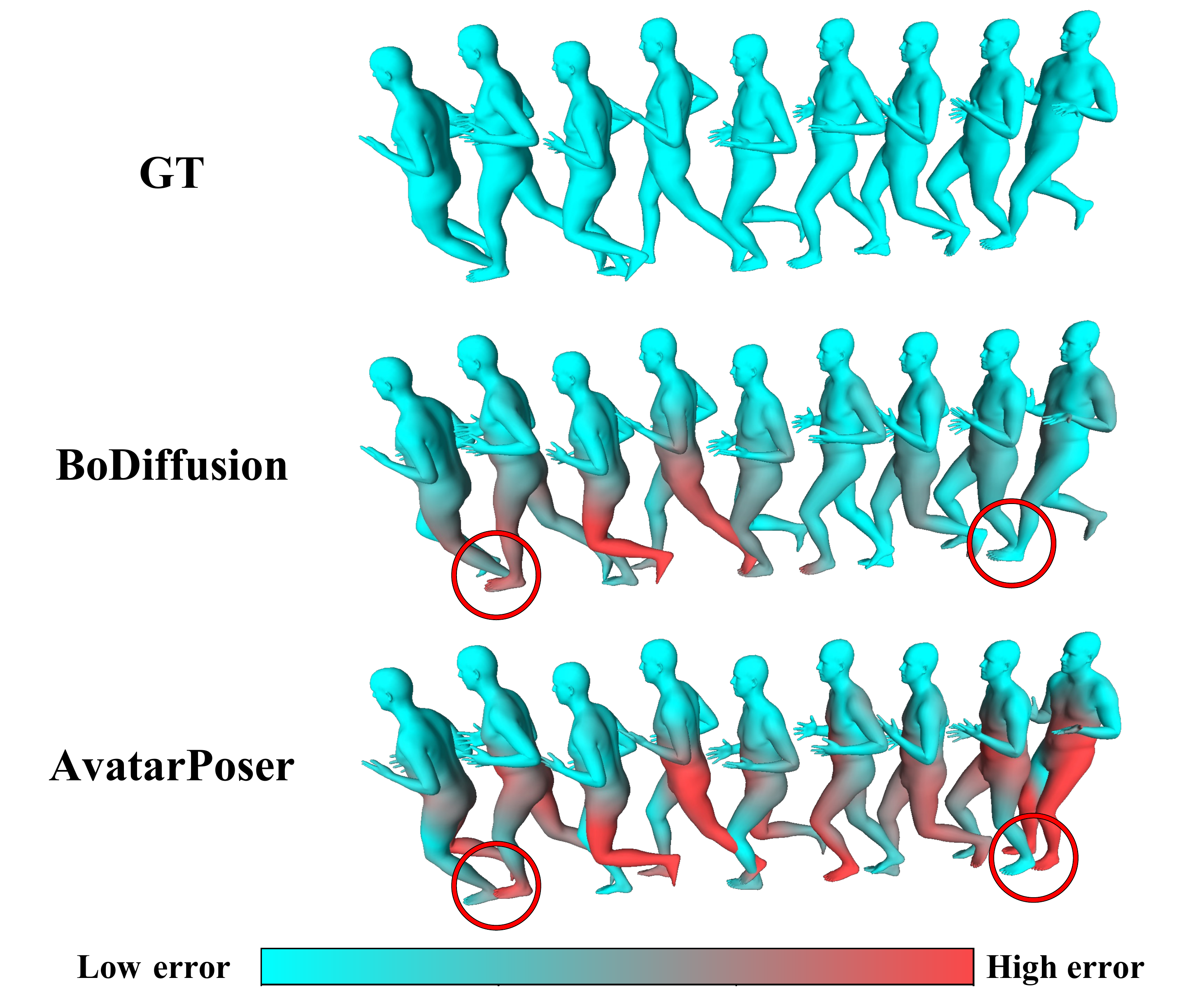}
    \caption{\textbf{Error Comparison.} Comparison of \MODELNAME\ and AvatarPoser~\cite{jiang2022avatarposer} with color coding as previously explained. Motions generated by \MODELNAME\ exhibit greater similarity to the ground truth and display fewer foot skating artifacts, as highlighted in the red circles. Specifically, the leg in contact with the ground should not slide, and \MODELNAME\ produces motion sequences that adhere more closely to this requirement.
    }
    \label{fig:qualitative_results_2}
\end{figure}

\paragraph{Probabilistic Framework.}
We formally define our conditional full-body motion synthesis task by using the formulation of Diffusion Models outlined in Sect.~\ref{sec:diffusion_background}.
Let $\x_t=x^{1:W}_t$, $\mathbf{s}=s^{1:W}$ for brevity.
We want to learn a conditional distribution of the full-body human motion sequences $\x_0$ defined as follows:

\begin{equation}\label{eq:marginal_cond_p_theta}
    p_{\theta}(\x_0| \mathbf{s}) = \int p_{\theta}(\x_{0:T} | \mathbf{s}) d \x_{1:T},
\end{equation}

\begin{equation}
    p_{\theta}(\x_{0:T}| \mathbf{s}) =  p(\x_T) \prod_{t=1}^T p_{\theta}(\x_{t-1} | \x_{t}, \mathbf{s}),
\end{equation}
where $p(\x_T) \sim \mathcal{N}(0, \mathbf{I})$ is a Gaussian noise.
In this case, we train a neural network $\theta$ to predict the mean $\mu_{\theta} (\x_t, t, \mathbf{s})$ and the variance $\Sigma_\theta(\x_t, t, \mathbf{s})$, similar to Eq.~\ref{eq:p_inverse}, but conditioned on sparse tracking signals $\mathbf{s}$. Thus, the simple loss from Eq.~\ref{eq:lsimple} then becomes:
\begin{equation}\label{eq:lsimple_motion}
    \mathcal{L}_{\text{simple}} = E_{\x_0 \sim q(\x_0), t \sim U\{1,T\}}||\epsilon - \epsilon_{\theta}(\x_t, t, \mathbf{s}) ||^2_2.
\end{equation}

\paragraph{Local Rotation Loss is Equivalent to the $\mathcal{L}_{\text{simple}}$.}
In Human Motion Synthesis, it is widespread~\cite{jiang2022avatarposer,dittadi2021full_VAE_HMD,aliakbarian2022flag,yang2021lobstr} to use the local rotation loss that minimizes the difference between the local joint rotations of the estimated poses and the ground truth.
Because of this standard practice, one can hypothesize whether learning $\epsilon_\theta$ (from Eq.~\ref{eq:lsimple}) is helpful for synthetic motion sequences. 
However, we found that optimizing $\epsilon_\theta$ is equivalent to directly minimizing the local rotation error.

\begin{lemma} \label{eq:lemma1}
Let $\mathcal{L}(x, x')=||x - x'||^2$ be the local rotation error loss between a motion sequence $x$ and $x'$ be an estimate of $x$. Then, optimizing the $\mathcal{L}_{\text{simple}}$ loss is equivalent to optimizing $\mathcal{L}$.
\end{lemma}
We provide the proof of Lemma \ref{eq:lemma1} in the Supplementary Material.

\subsection{\MODELNAME\ Architecture}\label{sec:bodiffusion_arch}
We draw inspiration from the diffusion models for image synthesis to design a model for learning the conditional distribution $p_{\theta}(x_0^{1:W}| s^{1:W})$ of the full-body motion sequences (cf. Eq.~\ref{eq:marginal_cond_p_theta}).
Specifically, we choose to leverage the novel Transformer backbone DiT~\cite{Peebles2022DiT} to build the \MODELNAME\ model because (i) it was shown to be superior for image synthesis task~\cite{Peebles2022DiT} compared to the frequently used UNet backbone \cite{Dhariwal2021DiffusionMB,ramesh2022hierarchical,rombach2022high}, and (ii) it is more naturally suited for modeling heterogeneous motion data.
Below, we provide a detailed description of our architecture and introduce a method that ensures the conditional generation of motion coherent with the provided sparse tracking signal $s^{1:W}$.

In order to leverage the Transformer's ability to handle long-term dependencies while maintaining temporal consistency, we format the input $x_t^{1:W}$, which represents joint rotations over time, as a time-sequence tensor and split it along the time dimension into tokens. We treat each pose $x^i_t$ as an individual token and combine the feature and joint dimensions into a $d$-dimensional vector, where $d=22\times6$ is the number of joints multiplied by the number of features. This strategy allows us to take advantage of the temporal information and efficiently process the motion sequence.

\begin{table*}[t!]
    \centering
    \footnotesize
     \resizebox{0.9\textwidth}{!}{%
    \begin{tabular}{lcccccccccccc}
         Method & Jitter & MPJVE & MPJPE & Hand PE & Upper PE & Lower PE & MPJRE & FCAcc $\uparrow$ \\
        \toprule
         Final IK* & - & 59.24 & 18.09 & - & - & - & 16.77 & -\\
         LoBSTr* & - & 44.97 & 9.02 & - & - & - & 10.69 & - \\
         VAE-HMD* & - & 37.99 & 6.83  & - & - & - & 4.11 & - \\
         AvatarPoser~\cite{jiang2022avatarposer} & 1.53 & 28.23 & 4.20 & 2.34 & 1.88 & 8.06 & 3.08 & 79.60\\
         AvatarPoser-Large~\cite{jiang2022avatarposer} & 1.17 & 23.98 & 3.71 & 2.20 & 1.68 & 7.09 & \textbf{2.70} & 82.30\\
         \MODELNAME\ (Ours) & \textbf{0.49} & \textbf{14.39}	& \textbf{3.63}	& \textbf{1.32}	& \textbf{1.53}	& \textbf{7.07}	& \textbf{2.70}	& \textbf{87.28}\\
        \bottomrule
    \end{tabular}
      } 
      \vspace{4pt}
    \caption{\textbf{Comparison with State-of-the-art Methods for Full-Body Human Pose Estimation.} Results on a subset of the AMASS dataset (CMU, BMLrub, and HDM05) for Jitter [km$/\text{s}^3$], MPJVE [cm$/$s], MPJPE [cm], Hand PE [cm], Upper PE [cm], Lower PE [cm], MPJRE [$\deg$], and FCAcc [\%] (balanced foot contact accuracy) metrics. AvatarPoser is retrained with 3 and 10 (Large) Transformer layers. The star (*) denotes the results reported in~\cite{jiang2022avatarposer}.}
    \label{tab:AvatarPoser_comp}
\end{table*}

We implement our \MODELNAME\ model by extending the DiT architecture of Peebles~\etal~\cite{Peebles2022DiT} with our novel conditioning scheme.
The DiT backbone architecture consists of a stack of encoder transformer layers that use Adaptive Layer Normalization (AdaLN). 
The AdaLN layers produce the scale and shift parameters from the timestep embedding vector to perform the normalization depending on the timestep $t$.
Peebles~\etal~\cite{Peebles2022DiT} input the class labels along with the time embedding to the AdaLN layers to perform class-conditioned image synthesis. 
However, we empirically demonstrate (see Sect.~\ref{sect:ablations}) that using the conditioning tracking signal $\mathbf{s}$ along with the time embedding $t$ in the AdaLN layers harms the performance of our \MODELNAME\ model because in this case, we disregard the time information.
Therefore, we propose a novel conditioning method that retains the temporal information and allows conditional synthesis coherent with the provided sparse tracking signal. 

\textbf{Conditioning on tracking signal.}
We use the 3-point tracking information of head and hands from HMDs to compute an enriched input conditioning $s^{1:W}$.
This conditioning $s^{1:W}$ has the shape $W \times d_s$, where $d_s=18\cdot3$ is the number of features (18) per joint multiplied by the number of tracked joints (3). We treat it as a sequence of individual tokens $s_i$ and apply a linear transformation (\emph{conditioning embedding} layer in Fig.~\ref{fig:pipeline}) to each of them, thus increasing the dimensionality of the tokens from $d_s$ to $d_{emb}=18\cdot22$. We observe that such higher-dimensional embedding enforces the model to pay more attention to the conditioning signal.
Next, we concatenate the input sequence tokens $x_t^i$ with the transformed conditioning tokens and input the result to the transformer backbone. 
By preserving the temporal structure of the tracking signal, we enable the model to efficiently learn the conditional distribution of motion where each pose in the synthesized sequence leverages the corresponding sparse tracking signal $s^i$.

\begin{table*}[t!]
    \centering
    \resizebox{0.8\textwidth}{!}{%
    \begin{tabular}{lcccccccc}
        Method & Jitter & MPJVE & MPJPE & Hand PE & Upper PE & Lower PE & MPJRE & FCAcc $\uparrow$ \\
        \toprule
         VAE-HMD (3p + pelvis)* & - & - &  7.45 & - & 3.75 & - & - & - \\
         VPoser-HMD (3p + pelvis)* & - & - & 6.74 & - & 1.69 & - & - & - \\
         HuMoR-HMD (3p + pelvis)* & - & - &  5.50 & - & 1.52 & - & - & - \\
         ProHMR-HMD (3p + pelvis)* & - & - &  5.22 & - & 1.64 & - & - & - \\
         FLAG~\cite{aliakbarian2022flag} (3p + pelvis)* & - &  - & \textbf{4.96} & - & \textbf{1.29} & - & - & -\\
         \midrule
         AvatarPoser~\cite{jiang2022avatarposer} (3p) & 1.11 & 	34.42 & 	6.32 & 	3.03 & 	2.56 & 	12.60	 & 4.64 & 	71.46\\
         \MODELNAME\  (Ours) (3p) & \textbf{0.35} & 	\textbf{21.37} & 5.78 & \textbf{1.94 }& 2.27 & \textbf{11.55} & \textbf{4.53} & \textbf{82.04}\\
        \bottomrule
    \end{tabular}
     } 
    \vspace{3pt}
    \caption{\textbf{Comparison against Generative-based Models. } Results reported on the held-out Transitions~\cite{AMASS:ICCV:2019} and HumanEVA~\cite{HEva_Sigal:IJCV:10b} subset of AMASS, following the protocol of FLAG~\cite{aliakbarian2022flag}, for Jitter [km$/\text{s}^3$], MPJVE [cm$/$s], MPJPE [cm], Hand PE [cm], Upper PE [cm], Lower PE [cm], MPJRE [$\deg$], and FCAcc [\%] metrics. We report the results after retraining AvatarPoser, and report  the same results as in~\cite{aliakbarian2022flag} for methods with a star (*).}
    \label{tab:FLAG_comp}
\end{table*}

\section{Experiments}

\textbf{Datasets.} We use the AMASS~\cite{AMASS:ICCV:2019} dataset for training and evaluating our models. AMASS is a large-scale dataset that merges 15 optical-marker-based MoCap datasets into a common framework with SMPL~\cite{SMPL:2015} model parameters. For our first set of experiments, we use the CMU~\cite{cmuWEB}, BMLrub~\cite{BMLrub}, and HDM05~\cite{MPI_HDM05} subsets for training and testing. We follow the same splits of AvatarPoser~\cite{jiang2022avatarposer} to achieve a fair comparison. For our second set of experiments, we evaluate the Transitions~\cite{AMASS:ICCV:2019} and HumanEVA~\cite{HEva_Sigal:IJCV:10b} subsets of AMASS and train on the remaining datasets following the protocol described in~\cite{aliakbarian2022flag}. 

\textbf{Evaluation Metrics.} We report four different types of metrics to evaluate our performance comprehensively. First, we report the velocity-related metrics Mean Per Joint Velocity Error [cm/s] (MPJVE), and Jitter error [km$/\text{s}^3$]~\cite{TransPoseSIGGRAPH2021} that measure the temporal coherence and the smoothness of the generated sequences. Second, we report the position-related metrics Mean Per Joint Position Error [cm] (MPJPE), Hand Position Error [cm] (Hand PE), Upper Body Position Error [cm] (Upper PE), and Lower Body Position Error (Lower PE). The third set is rotation-related metrics, including the Mean Per Joint Rotation Error [$\deg$] (MPJRE). Finally, we devise a metric based on Foot Contact (FC) to measure if the predicted body has a realistic movement of the feet. To calculate this metric for every pair of instances in a sequence, we determine if there is contact between the four joints of the feet and the ground by calculating the velocity of the joints and checking whether it is under a pre-defined threshold or not, following~\cite{tevet2022human}. Afterward, we calculate the accuracy between the predicted and the ground-truth FC. Since the ratio of foot contact vs. foot in the air is meager, we calculate a balanced accuracy (FCAcc).

\begin{figure}
    \centering
    \includegraphics[width=0.8\linewidth]{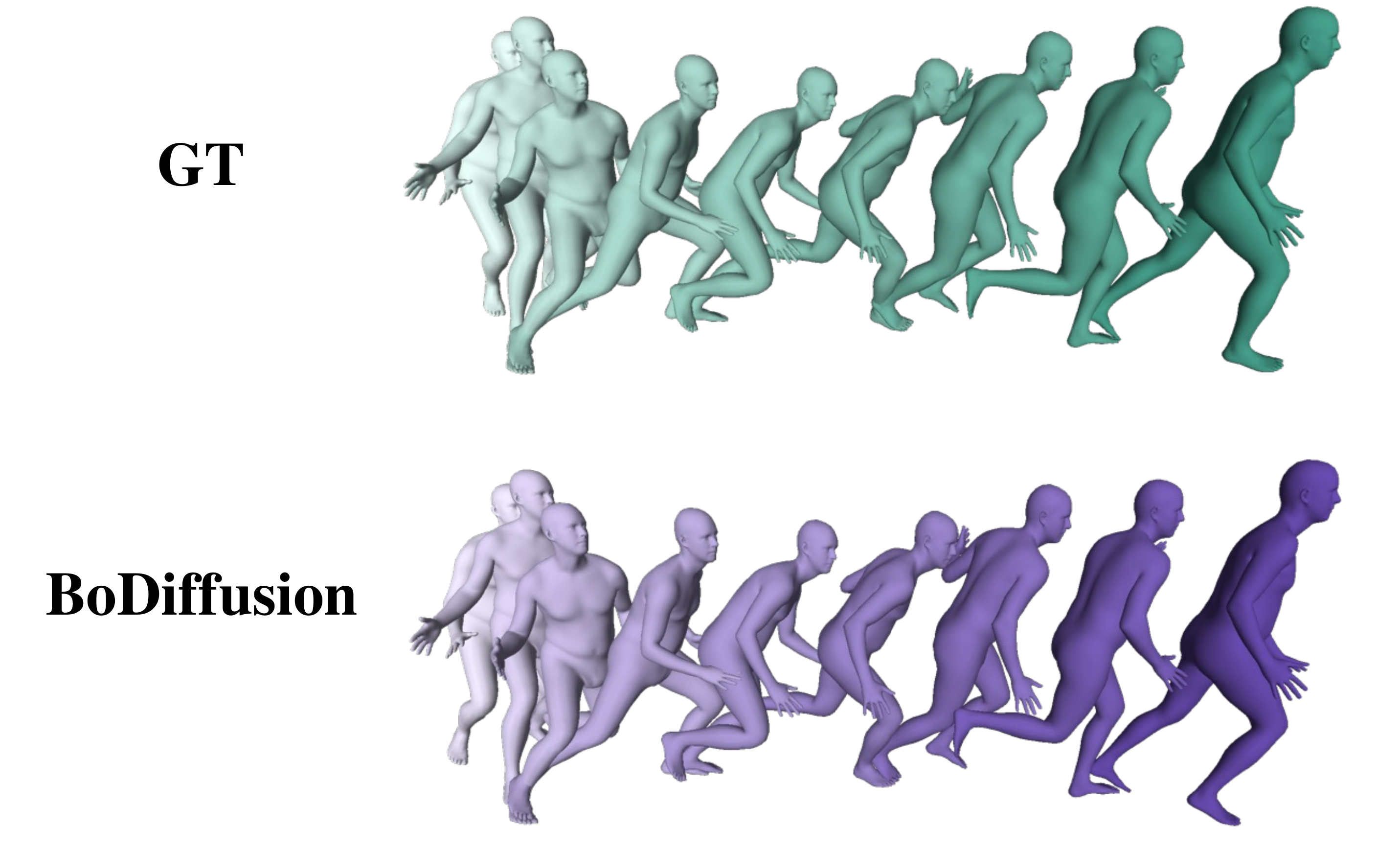}
    \caption{\textbf{Full-Sequence Generation.} Example prediction of \MODELNAME\ compared against the ground-truth sequence. Our method can generate realistic motions faithful to the ground truth. Color gradient represents time flow, whereas lighter colors denote the past.}
    \label{fig:sequence_comparison}
\end{figure}

\textbf{Implementation Details.} Similar to~\cite{jiang2022avatarposer}, we set window size $W=41$. 
Our Transformer backbone consists of $12$ DiT blocks~\cite{Peebles2022DiT}.
Before feeding to the backbone, the input tokens are projected to the hidden dimension $emb=384$, as shown in Fig.~\ref{fig:pipeline}.
Finally, we project the output of the last DiT block back to the human body pose space of shape ${41 \times 6 \cdot 22}$, representing the 6D rotations for 22 body joints.
During training, we use $ \lambda_{\text{vlb}} = 1.0$, and define $t$ to vary between $[1, T]$, where $T=1000$ corresponds to a pure Gaussian distribution.
At inference, we start from pure Gaussian noise, and we use DDIM sampling~\cite{song2020ddim} with $50$ steps. 
We set the variance $\Sigma_\theta$ of the reverse noise to zero.
This configuration turns the model into a deterministic mapping from Gaussian noise to motions, allowing it to do much fewer denoising steps without degrading the quality of synthesized motions.

We use AdamW optimizer~\cite{loshchilov2018fixing} with a learning rate of $1e-4$, batch size of $256$, without weight decay. Our model has 22M parameters and is trained for 1.5 days on four NVIDIA Quadro RTX 8000. More implementation details are in the Supplementary Material (Sec.~\ref{sec:imp_details_supp}).

Our approach has no limitations concerning the length of the generated sequences. We can synthesize motions of arbitrary length by applying \MODELNAME\ in an autoregressive manner using a sliding window over the input data. We refer the reader to the Supplementary Material for more explanation of our inference-time protocol (see Sec.~\ref{sec:inference_supp}). 

\subsection{Results}\label{sec:results}
We compare \MODELNAME~with AvatarPoser~\cite{jiang2022avatarposer} and FLAG~\cite{aliakbarian2022flag} following their experimental setups. For AvatarPoser in Table~\ref{tab:AvatarPoser_comp}, we use the official source code to retrain the standard version with 3 Transformer layers. Furthermore, to ensure a fair comparison with \MODELNAME, we train a scaled-up version of AvatarPoser (AvatarPoser-Large) with 10 layers, 8 attention heads, and an embedding dimension of 384. Since the other state-of-the-art methods do not provide public source codes, we compare them against the results reported in each of the previous papers.  

Table~\ref{tab:AvatarPoser_comp} shows that \MODELNAME\ outperforms the state-of-the-art approaches in all metrics on the test subset of the AMASS dataset (CMU, BMLrub, and HDM05). 
Since we enforce the temporal consistency in \MODELNAME\ by leveraging the novel conditioning scheme and learning to generate sequences of poses instead of individual poses, our method generates smoother and more accurate motions. 
This is demonstrated by our quantitative results in Tab.~\ref{tab:AvatarPoser_comp}. 
We observe a significant improvement in the quality of generated motions by leveraging the \MODELNAME~model.
Thus, we are able to decrease the MPJVE by a margin of 9.59 cm$/$s and the Jitter error by 0.68 km$/\text{s}^3$, compared to AvatarPoser-Large. 
Fig.~\ref{fig:qualitative_results_2} shows that motions generated by \MODELNAME\ exhibit more significant similarity to the ground truth across all the sequence frames and display fewer foot-skating artifacts compared to AvatarPoser, which struggles to maintain coherence throughout the sequence and severely suffers from foot skating.
Furthermore, we empirically demonstrate that our method successfully learns a manifold of plausible human poses while maintaining temporal coherence.
In practice, we are given the global position of the hands and head as the conditioning; thus, it is expected to have a lower error on these joints, while the conditioning does not uniquely define the configuration of legs and should be synthesized.
However, Fig.~\ref{fig:qualitative_results_1}, \ref{fig:qualitative_results_2}, \ref{fig:sequence_comparison} show that \MODELNAME\ produces plausible poses not only for the upper body but for the lower body as well, in contrast to the state-of-the-art Transformer-based AvatarPoser method.

Fig.~\ref{fig:qualitative_results_1} qualitatively shows the improvement of our method in positional errors. In particular, our method predicts lower body configurations that resemble the ground truth more than AvatarPoser. These results support the effectiveness of our conditioning scheme for guiding the generation towards realistic movements that are in close proximity to the ground-truth sequences.

Furthermore, our method achieves a better performance in the Foot Contact Accuracy metric (FCAcc), as shown in Table~\ref{tab:AvatarPoser_comp} and the feet movements in Fig.~\ref{fig:sequence_comparison}. Thus, the iterative nature of the DDPMs, along with our spatio-temporal conditioning scheme, allows us to generate sequences with high fidelity even at the feet, which are the furthest from the input sparse tracking signals. 

Table~\ref{tab:AvatarPoser_comp} shows the performance of a larger version of AvatarPoser-Large compared to ours.
In particular, we demonstrate that enlarging this model increases its motion capture capacity to the point where it reaches more competitive results.  
By definition, this experiment also demonstrates that using more complex methods leads to better performance. 
However, \MODELNAME\ depicts a better trade-off between the performance and computational complexity than state-of-the-art methods.
Since \MODELNAME\ can take advantage of DiT, our approach will further improve in the measure that foundation models reach better results. 

\begin{table}[]
\resizebox{\linewidth}{!}{%
\centering
\begin{tabular}{lccccccc@{}}
Method & Jitter & MPJVE & MPJPE & MPJRE \\ 
\toprule
\MODELNAME\ (Token input cond) & \textbf{0.49} & \textbf{14.39}	& 3.63 	& 2.70	\\
Timestep cond & 1.38 & 52.78 & 7.19 & 4.00 \\
Token input + Timestep cond & 0.59 & 16.22 & 3.60  & \textbf{2.60}  \\
with stochasticity & 0.53 &15.37 & \textbf{3.53} &2.67 \\
\midrule
Window size W=1 & 19.71 & 174.9 & 4.77 & 3.13  \\
Shuffled sequences &  108.42 & 935.69 & 17.13 & 7.10\\
\bottomrule
\end{tabular}%
} 
\vspace{2pt}
\caption{\textbf{Design Ablations.} \textbf{Up:} We ablate our training scheme by varying the conditioning approach. At inference, we demonstrate that controlling the stochasticity smoothens our predictions. \textbf{Down:} We assess the importance of including temporal context.} 
\label{tab:ablations}
\end{table}

\begin{table}[]
\resizebox{\linewidth}{!}{%
\centering
\begin{tabular}{lcccc@{}}
Method & Jitter & MPJVE & MPJPE & MPJRE \\ 
\toprule
UNet w/o diffusion & 1.44 & 33.35 & 4.36 & 2.81 \\
Transformer w/o diffusion & 1.27 & 27.62 & 3.92 & 2.60 \\
\midrule
\MODELNAME-UNet & 1.24 & 20.65 & \textbf{3.63} & \textbf{2.48} \\
\MODELNAME-Transformer (Ours) & \textbf{0.49} & \textbf{14.39}	& \textbf{3.63}	& 2.70\\
\bottomrule
\end{tabular}%
 } 
 \vspace{2pt}
\caption{\textbf{Architecture Ablations.} We evaluate the relevance of using DiT as our backbone. We also assess the effectiveness of the denoising power of our DDPM by comparing it against the backbones without diffusion.} 
\label{tab:ablations_arch}
\end{table}

\begin{table}[]
\centering
\resizebox{0.75\linewidth}{!}{%

\begin{tabular}{ccccc@{}}
DDIM steps & Jitter & MPJVE & MPJPE & MPJRE \\ 
\toprule
10 & 0.56 & 16.16 & 3.89 & 2.84\\
20 & 0.52 & 15.05 & 3.72 & 2.75\\
30 & 0.51 & 14.75 & 3.66 & 2.73\\
40 & 0.49 & 14.55 & 3.64 & 2.71\\
50 & 0.49 & 14.39 & 3.63 & 2.70\\
100 & 0.48 & 14.12 & 3.44 & 2.59\\
\bottomrule
\end{tabular}%
 } 
 \vspace{2pt}
\caption{\textbf{Ablation of inference sampling steps.} During inference, we use DDIM sampling with 50 steps. Note that the performance improves when there are more sampling steps. }
\label{tab:iter}
\end{table}

Table~\ref{tab:FLAG_comp} shows the quantitative comparison between~\MODELNAME\ and other generative-based state-of-the-art approaches for the Transitions~\cite{AMASS:ICCV:2019} and HumanEVA~\cite{HEva_Sigal:IJCV:10b} subsets of AMASS. AvatarPoser is included for reference. On the one hand, even though we only train with three sparse inputs, we have competitive results regarding an overall positional error (MPJPE) and upper body positional error (Upper PE) with the methods that also use the pelvis information. Our DDPM-based method outperforms the VAE-based approaches VAE-HMD and VPoser-HMD and has comparable results with the conditional flow-based models ProHMR-HMD and FLAG. On the other hand, our \MODELNAME\ has a better performance than AvatarPoser in all the metrics, with a significant improvement in the velocity-related metrics MPJVE and Jitter. Please refer to the Supplementary for additional qualitative results. 

\subsection{Ablation Experiments}\label{sect:ablations}


We conduct ablation experiments to assess the effect of the different components of our method on the smoothness and temporal consistency of the generated sequences. 
In Table~\ref{tab:ablations}, we report the experiments corresponding to the conditioning scheme, the stochastic component at inference, and the relevance of temporal context.
Firstly, we compare the effect of using different conditioning schemes. Our method receives the conditioning by concatenating the input tokens (Token input cond). Thus, the conditioning keeps time-dependent information, allowing us smoother predictions, as the low Jitter and MPJVE values show. In contrast, applying the condition through the timestep embedding (Timestep cond) results in a compression towards a time-agnostic vector embedding. Table~\ref{tab:ablations} shows that using this time-agnostic embedding solely as conditioning results in detrimental performance for the method. Furthermore, using both the token input and the timestep conditioning still results in less smooth sequences and is less consistent than using only the token input conditioning scheme. 

Secondly, we implemente a purely stochastic inference scheme (w/ stochasticity), finding out that, even when the rotational and positional errors decrease slightly, having extra control over the randomness is beneficial, especially for the smoothness of the sequences, as shown by the decrease in MPJVE and Jitter.  
Thirdly, we evaluate the importance of having temporal consistency by using a sliding window of size one during training (Window size W=1) and randomly sorting the sequence at inference time (Unordered sequence). As expected, the MPJVE and Jitter errors increase significantly, and all the other metrics also increase by some proportion. Therefore, these experiments confirm the relevance of enforcing temporal consistency. 

Table~\ref{tab:ablations_arch} presents the impact of different architectural choices on the performance of our proposed model.
First, to validate the effectiveness of using DiT as our backbone (BoDiffusion-Transformer), we compare it against UNet (BoDiffusion-UNet), which has traditionally been used as a backbone for diffusion models~\cite{Dhariwal2021DiffusionMB,rombach2022high}. Table ~\ref{tab:ablations_arch} indicates that the Transformer outperforms UNet in all the metrics, even when diffusion processes are not involved. Additionally, when incorporating our diffusion framework on top of both backbones, significant improvements are observed in the temporal consistency and quality of the generated sequences. It is important to note that while replacing the DiT backbone with UNet leads to a slight decrease ($0.2\degree$) in rotation error, it is accompanied by a significant increase in Jitter and Velocity errors. Thus, these ablation experiments demonstrate the complementarity of using a transformer-based backbone in a diffusion framework, resulting in smoother and more accurate predictions. 

Table \ref{tab:iter} shows the ablation experiment using different sampling steps for DDIM at inference time. Increasing the sampling steps improves the performance of our method, proving the importance of the iterative nature of DDPMs. However, more steps require more computational capacity. Thus, we select 50 DDIM steps for an appropriate trade-off between performance and complexity.

\section{Conclusion}

In this work, we present \MODELNAME, a Diffusion model for conditional motion synthesis inspired by effective architectures from the image synthesis field. Our model leverages the stochastic nature of DDPMs to produce realistic avatars based on sparse tracking signals of the hands and head. 
\MODELNAME\ uses a novel spatio-temporal conditioning scheme and enables motion synthesis with significantly reduced jittering artifacts, especially on lower bodies.
Our results outperform state-of-the-art methods on traditional metrics, and we propose a new evaluation metric to fully demonstrate \MODELNAME 's capabilities. 

\subsection*{Acknowledgements}
Research reported in this publication was supported by the Agence Nationale pour la Recherche (ANR) under award number ANR-19-CHIA-0017.

{\small
\bibliographystyle{ieee_fullname}
\bibliography{egbib}
}

\clearpage

\appendix

\twocolumn[{%
\renewcommand\twocolumn[1][]{#1}%
\maketitle
\begin{center}
    \centering
    \includegraphics[width=0.9\linewidth]{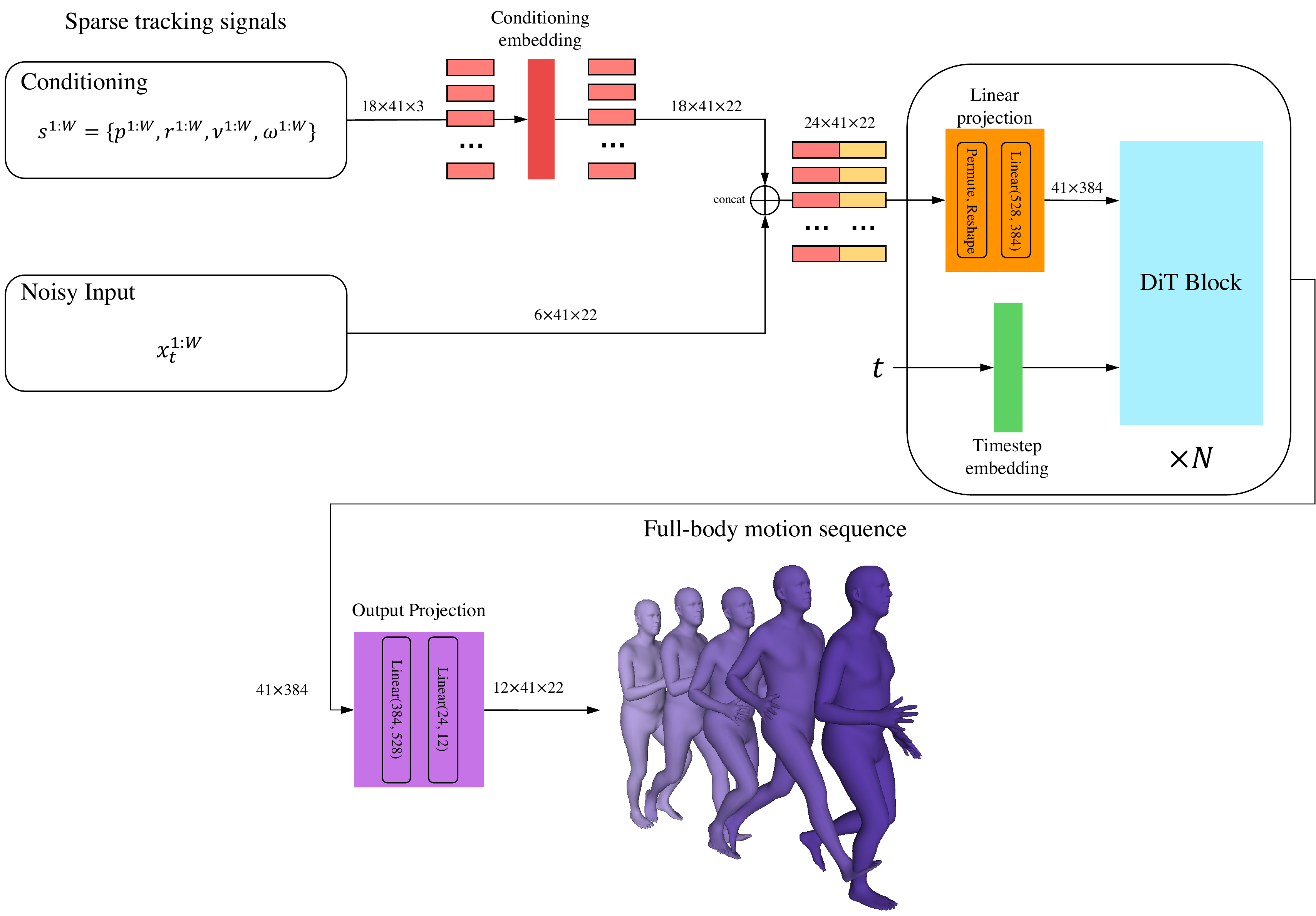}
    \captionof{figure}{
    \textbf{Complete Overview of~\MODELNAME.} Our conditional model takes full advantage of sparse information since we calculate relevant features in the conditioning pathway at the top (red block). In the case of the noisy input, we do not need any projection to match the sizes from the conditioning pathway. After concatenating both pathways, we organize the tensors' dimensions and perform an additional projection. The linear projection changes the tensor dimensions to the embedding dimension for the DiT blocks. After denoising by the DiT, we perform a final projection to the original space of full body motions (purple block). The output estimates $\epsilon_{\theta}$ and $\Sigma_{\theta}$ that are used to compute the local rotations $x_{t-1}^{1:W}$ by sampling from $\mathcal{N}(x_{t-1}; \mu_{\theta}, \Sigma_\theta)$.  Here $W=41$ is the temporal window size, conditioning signal $s^{1:W}$ contains 18D features for the three tracked joints (head and hands), and $x_t^{1:W}$ is the noisy local 6D rotations for $22$ body joints. 
    $\oplus$ is the operation of concatenation along the channels' dimension. The numbers next to the arrows denote the input and output dimensions for the corresponding blocks. 
    }
    \label{fig:overview_dit}
    \vspace{2cm}
\end{center}
}]

\section{Implementation Details}

We build upon the SMPL~\cite{SMPL:2015,MANO:SIGGRAPHASIA:2017,SMPL-X:2019} parametric model that uses local rotations in axis-angle representation to produce a full-body pose. 
Our model predicts local rotations in 6D representation that are then converted to axis-angle representation to be used in the body model from SMPL.
As in~\cite{jiang2022avatarposer}, we use a neutral body model corresponding to the average body model between women and men.
We do not apply normalization to the conditioning signal $s^{1:W}$ before inputting it into the model.

\subsection{Architecture}\label{sec:imp_details_supp}
\begin{figure}[t]
    \centering
    \includegraphics[width=0.82\linewidth]{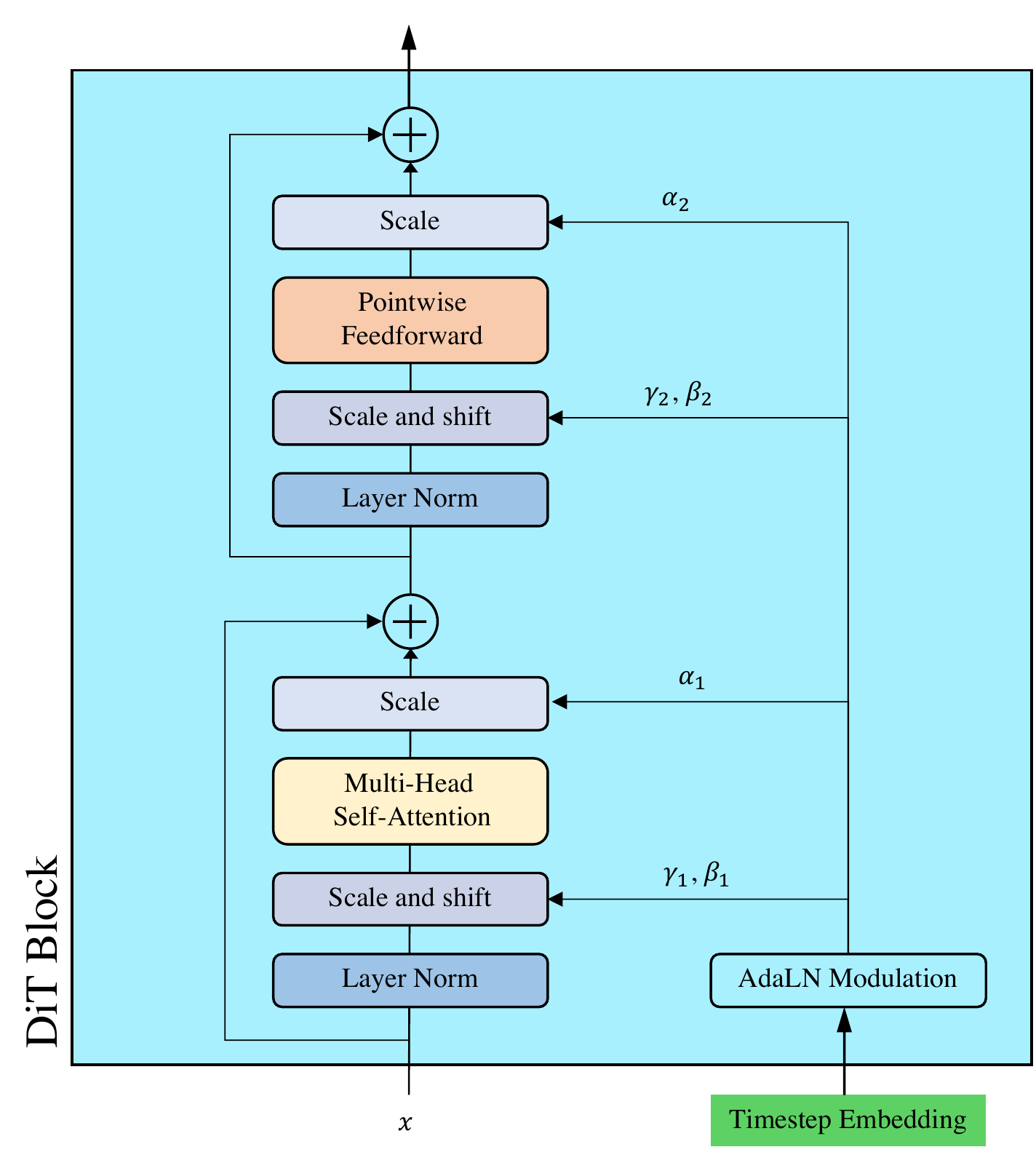}
    \caption{\textbf{\MODELNAME\ Architecture.} Our denoising model is built using multiple DiT blocks. Here we show the details of a single DiT block. 
    }
    \label{fig:dit}
\end{figure}

In Figures~\ref{fig:overview_dit} and~\ref{fig:dit}, we provide further information on the architecture of our model and technical details.
In Figure~\ref{fig:overview_dit}, we show the projection of the input condition (red block), which corresponds to the joint positions $p^{1:W}$, rotations $r^{1:W}$, linear velocities $v^{1:W}$, and angular velocities $\omega^{1:W}$ in the global coordinate frame. 
This projection aims to change the feature dimension of the conditioning 
input.
It is worth saying that the input $x_t$ corresponds to a noisy input at time $t$.
After denoising with the DiT, we perform a final projection (Figure~\ref{fig:overview_dit}, in purple) to map back into the space of motions represented by 6D local rotations of joints.
We return a $12$-channel tensor which contains predictions of $\epsilon_{\theta}$ and $\Sigma_{\theta}$ ($6$ channels each) that are used to compute losses $\mathcal{L}_{\text{simple}}$ and $\mathcal{L}_{\text{vlb}}$.

Figure~\ref{fig:dit} presents a detailed scheme of our DiT architecture for the denoising process. 
The DiT network starts with a Layer Normalization followed by an adaptive normalization that uses the timestep embedding. 
This adaptive normalization consists of an MLP that learns regression values that come from the embedding vectors of the timestep instead of learning the modulation parameters $\gamma$ and $\beta$ parameters from the data.  
Afterward, we use six attention heads in the self-attention and perform one more scaling from which values come from the adaptive normalization. 
We apply a residual connection between the scaling's input and output.
Then, we repeat the normalization stages, but instead of having another attention mechanism, we use the typical point-wise feedforward. 
In the end, we finish with another residual connection, which is a summation. 
We follow~\cite{Peebles2022DiT,song2020improved} to compute the timestep embedding.

\subsection{Inference}\label{sec:inference_supp}
At inference time, we use DDIM~\cite{song2020ddim} with $50$ iterations and remove the stochasticity during sampling from the distribution by setting the variance $\Sigma_{\theta}$ to zero.
Using a sliding window, we use the temporal window size $W=41$ and apply \MODELNAME\ to the input tracking signal from HMD and hand controllers. While during online inference, one would apply our model using a sliding window with stride $1$, for the sake of faster inference on AMASS dataset, we apply our model using a stride of $20$ frames. We did not observe the degradation of generations' quality when we increased the stride.

\subsection{Inference Speed}
A forward pass with $W =41$  takes 0.021 secs for our method and 0.003 secs for AvatarPoser. At inference, we do 50 forward passes that amount to 1.046 secs. Our method is not optimized for speed yet because our goal was to prove that DDPMs can generate high-quality motions. Future work has a huge potential for making DDPM’s inference faster by more efficient sampling, reducing the number of layers and channels, and using quantization.

\subsection{UNet Architecture for Ablation}
\begin{figure}
    \centering
    \includegraphics[width=0.7\linewidth]{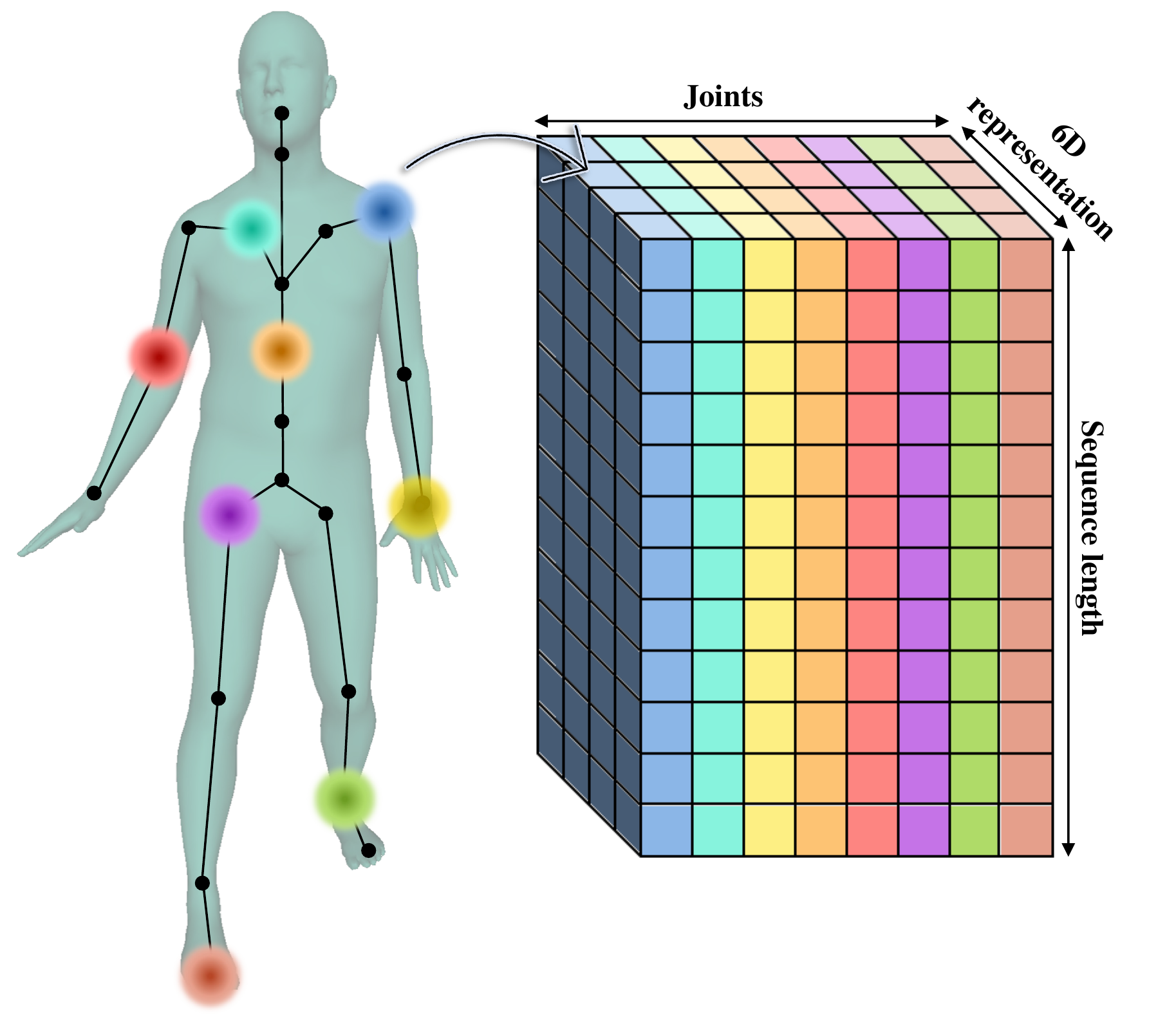}
    \caption{\textbf{Input tensor representation for UNet network.} 
    We represent the motion sequence as a 3D tensor in which the channels correspond to the 6D rotation, the height to the time sequence, and the width to the joints. This representation is analogous to the image data input tensors. In this way, we can reuse the convolutional architectures of the denoising UNet for 3D body pose estimation.}
    \label{fig:input_tensor}
\end{figure}

To ablate the architecture, we also implemented a version of \MODELNAME\ using the popular DDPM UNet backbone~\cite{Dhariwal2021DiffusionMB} designed for image data and not for motions, the overview of this architecture is shown in Fig.~\ref{fig:initial_proj}. 
We followed~\cite{Dhariwal2021DiffusionMB} for the architecture's hyper-parameter selection. 
In our case, we modified the ImageNet 128-channel architecture but changed the base number of channels from 256 to 64.
Furthermore, we kept the same hyper-parameters for the feature dimension multiplication.
Considering the motion represented as a sequence of poses $x^{1:W}$, we can treat it as a ``structured" image tensor (as shown in Fig.~\ref{fig:input_tensor}), such that spatial dimensions (height and width for image) are replaced by ``time" and ``joints" dimensions and channels are replaced by joint features (in our case it is 6D rotations). 
A ``structured" image in this context means that each pixel of the image represents a single joint located in the kinematic tree along one axis and time along another axis of the tensor. 
Due to the sufficient depth of the network and a self-attention block in the middle, the effective receptive field of the deepest convolutional layers covers the entire ``structured" image.


\section{Additional Ablation Experiment}
Table~\ref{tab:supp_ablations} demonstrates that our window size is optimal for this task. 
First, we empirically show that removing the temporal information from the input leads to high jitter and velocity errors. 
In practice, using single frames is not enough to enforce temporal consistency, thus making it harder to understand the full-body movement.
Therefore, even when the positional and rotational errors are not extremely high compared with our model, the jitter and velocity errors increase considerably, thus misspending the long-range analysis capacity of Transformers.
Secondly, we vary the number of input sequences to demonstrate the importance of enforcing temporal consistency.
Since our window size is $41$, we choose half and double the number of input sequences to assess the benefit of increasing or decreasing the temporal information.
As expected, increasing the window size to $81$ results in having more temporal coherence, thus decreasing the jitter and velocity errors.
However, increasing the input window size also increases the computational cost of training from 1.5 days to almost 3 days. 
In contrast, reducing the window size to $21$ leads to harnessing the smoothness of the motion.
It is worth mentioning that even when the jitter and velocity errors are affected by different window sizes, our method performs the best in terms of positional and rotational errors.


\begin{table}[]
\resizebox{\linewidth}{!}{%
\centering
\begin{tabular}{lccccccc@{}}
Method & Jitter & MPJVE & MPJPE & MPJRE \\ 
\toprule

Window size $W=1$ & 19.71 & 174.9 & 4.77 & 3.13  \\
Window size $W=21$ & 0.53 & 16.09 & 3.96 & 2.86  \\
\MODELNAME\ ($W=41$) & 0.49 & 14.39	& \textbf{3.63} & \textbf{2.70}	\\
Window size $W=81$ & \textbf{0.46} & \textbf{13.69} & 3.77 & 2.86  \\

\bottomrule
\end{tabular}%
} 
\vspace{2pt}
\caption{\textbf{Window Size Ablation.} We evaluate the importance of including more or less temporal context. We report Jitter [km$/\text{s}^3$], MPJVE [cm$/$s], MPJPE [cm], and MPJRE [$\deg$].}
\label{tab:supp_ablations}
\end{table}

\section{Additional Qualitative Evaluation}

\begin{figure*}[t]
    \centering
    \vspace{-5pt}
    \includegraphics[width=\linewidth, clip]{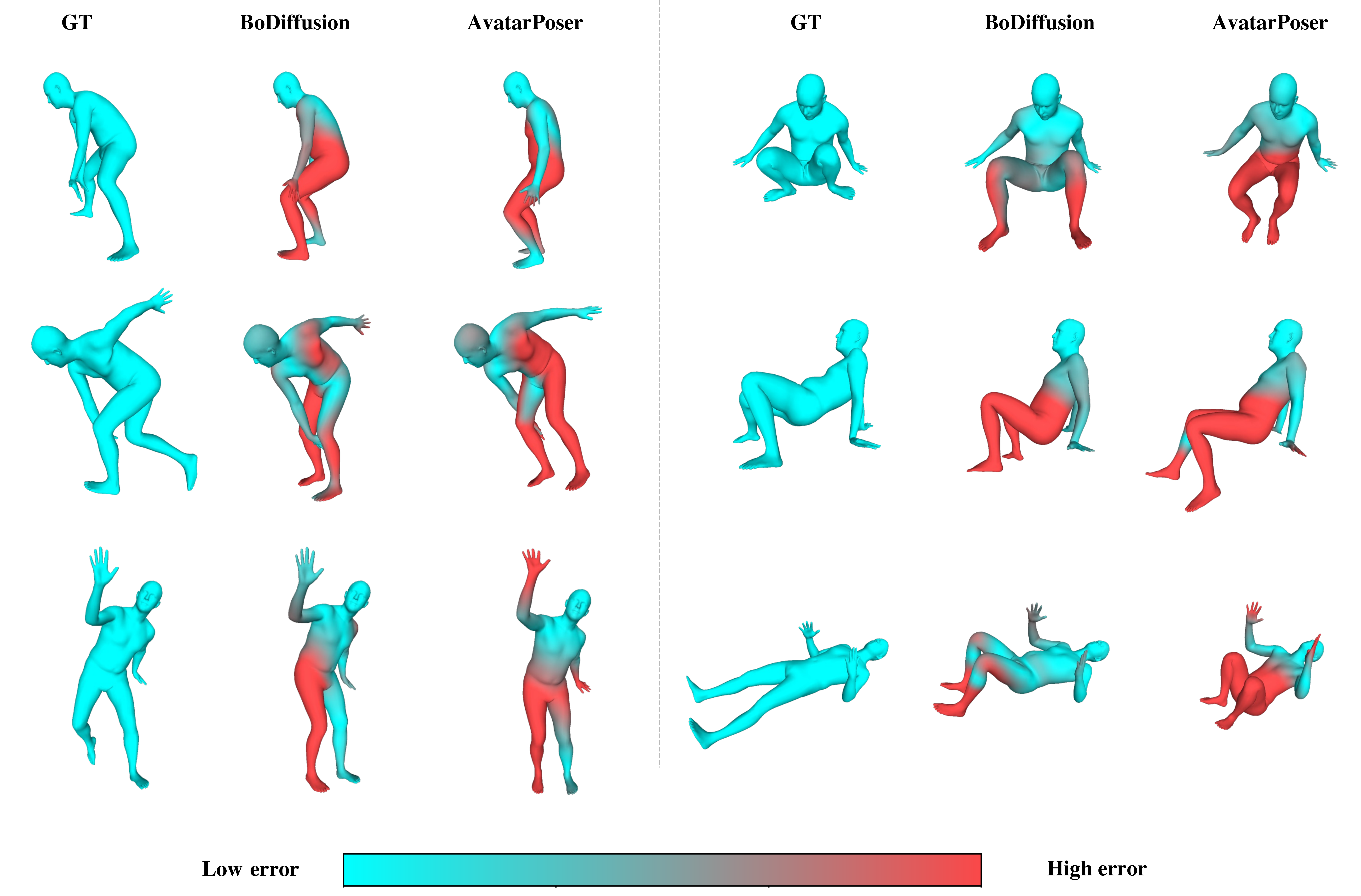}
    \vspace{-18pt}
    \caption{\textbf{Performance on unconventional poses.} We compare single poses predicted by using \MODELNAME~and AvatarPoser~\cite{jiang2022avatarposer}. The poses were extracted from sequences of the CMU, BMLRub and HDM5 datasets. Mesh colors denote absolute positional error. Note how our method can predict  plausible poses even for uncommon movements like crouching or lying down.}
    \label{fig:qualitative_results_indv_ap}
\end{figure*}
\begin{figure*}[t]
    \centering
    \vspace{-5pt}
    \includegraphics[width=\linewidth, clip]{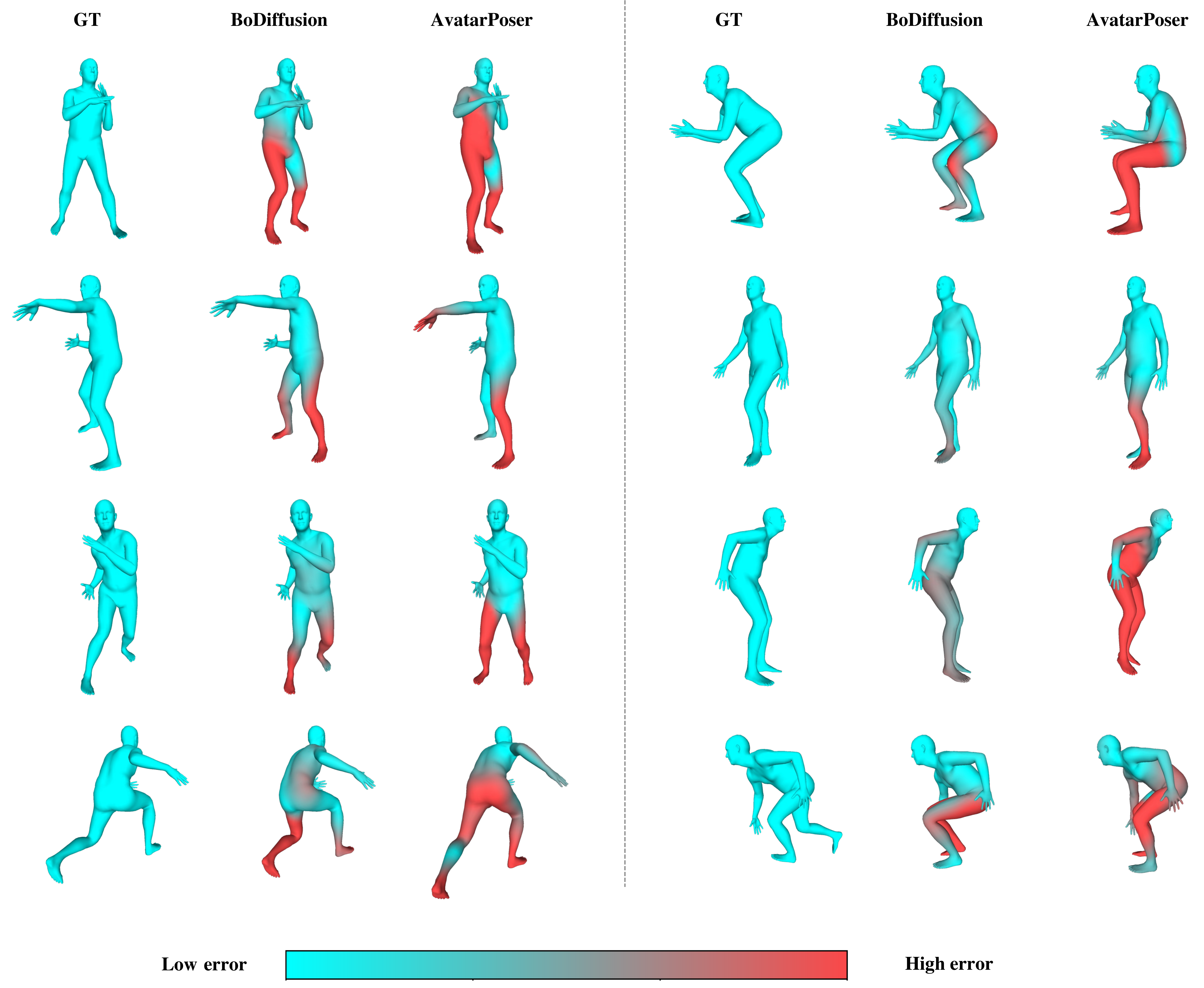}
    \vspace{-18pt}
    \caption{\textbf{Error visualization on individual poses.} We compare \MODELNAME~and AvatarPoser~\cite{jiang2022avatarposer} on sequences from the Transitions~\cite{AMASS:ICCV:2019} and HumanEVA~\cite{HEva_Sigal:IJCV:10b} datasets. Note how our method can predict poses with higher fidelity to the ground truth. In contrast, AvatarPoser struggles to predict accurate lower-body configurations.}
    \label{fig:qualitative_results_idv_flag}
\end{figure*}
\begin{figure*}[h]
    \centering
    \includegraphics[width=0.98\linewidth, clip]{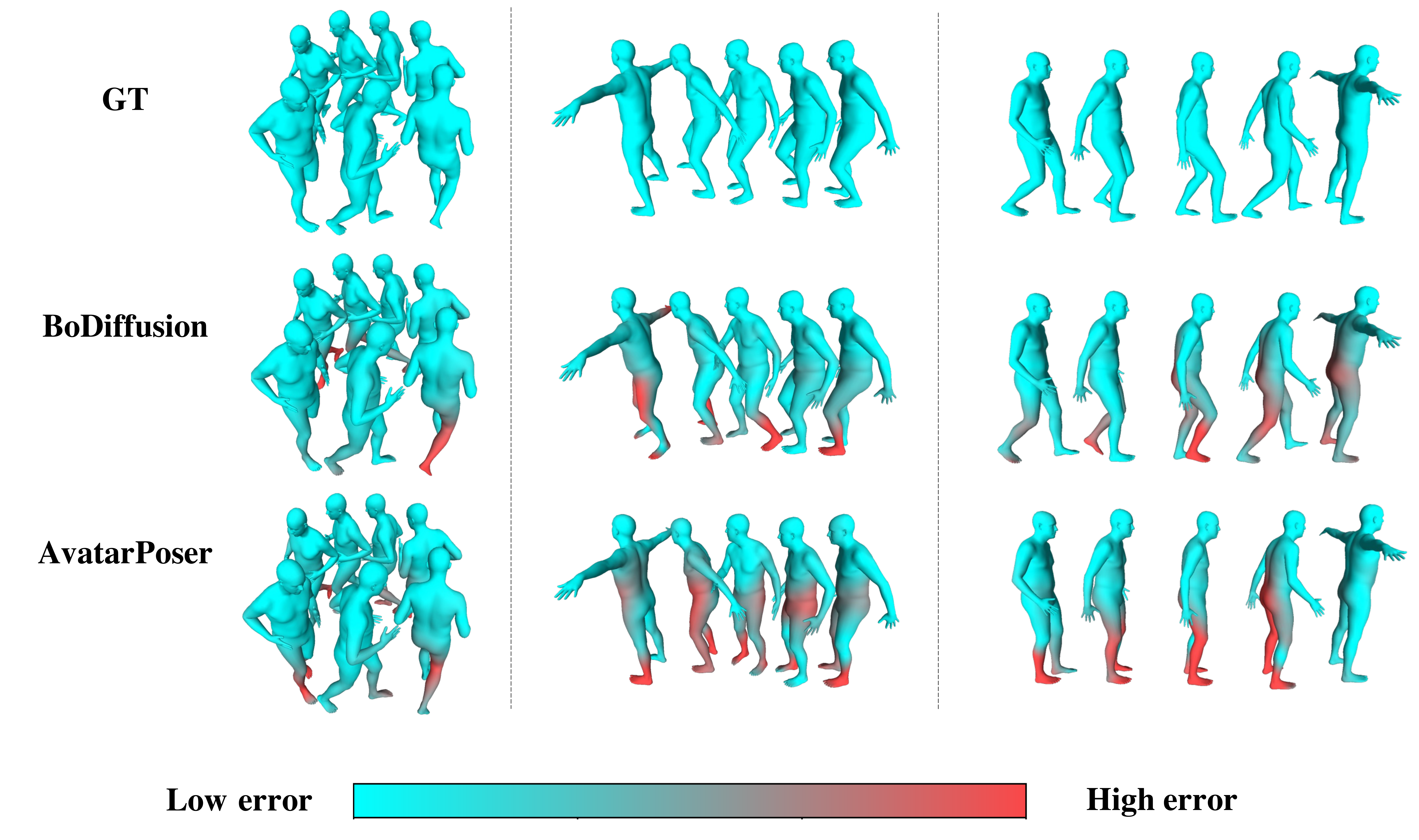}
    \vspace{-12pt}
    \caption{\textbf{Error visualization on sequences.} We compare predicted motions of \MODELNAME\ and AvatarPoser~\cite{jiang2022avatarposer} on the test sequences from the Transitions~\cite{AMASS:ICCV:2019} and HumanEVA~\cite{HEva_Sigal:IJCV:10b} datasets. Notice that the motions generated by \MODELNAME\ look more natural and demonstrate better temporal consistency. On the contrary, methods like AvatarPoser struggle to maintain coherence throughout the frames regarding aspects like foot sliding (third sequence).}
    \label{fig:qualitative_results_seq_flag}
\end{figure*}

Figure~\ref{fig:qualitative_results_indv_ap} shows additional qualitative results for~\MODELNAME~ and AvatarPoser~\cite{jiang2022avatarposer} on the CMU~\cite{cmuWEB}, BMLrub~\cite{BMLrub}, and HDM05~\cite{MPI_HDM05} test sets. Notice that our method can generate poses close to the ground truth even when the actions are unusual. For instance, column 2 depicts poses of a person doing movements very close to the ground. Our method is able to use the sparse tracking input for such uncommon motions and predict plausible body configurations that are faithful to the ground truth. In contrast, AvatarPoser struggles with creating accurate poses when seeing an uncommon motion.

Figures \ref{fig:qualitative_results_idv_flag} and \ref{fig:qualitative_results_seq_flag} show qualitative results on  the Transitions~\cite{AMASS:ICCV:2019} and HumanEVA~\cite{HEva_Sigal:IJCV:10b} test sets predicted with \MODELNAME~and AvatarPoser~\cite{jiang2022avatarposer}. First, note that \MODELNAME~ generates individual poses with a high fidelity in the upper-body configuration and a plausible lower-body configuration. Second, \ref{fig:qualitative_results_seq_flag} shows that \MODELNAME~ captures more details of the position of the feet and avoids foot sliding, unlike AvatarPoser. 

To fully appreciate the high quality of the motions generated by our approach, we suggest the reader watch the video attached to this supplementary material. The video demonstrates that \MODELNAME\ synthesizes more accurate motions with substantially less jitter than AvatarPoser~\cite{jiang2022avatarposer}.

\section{Local Rotation Loss}

Due to the properties of Gaussian distributions, Ho~\etal~\cite{ho2020denoising} showed that we can directly calculate $\x_t$ from $\x_0$ by sampling:
\begin{equation}\label{eq:x_t_supp}
    \x_t = \sqrt{\bar{\alpha}_t} \x_0 + \sqrt{1 - \bar{\alpha}_t} \epsilon,
\end{equation}
and the following simple loss function can be used for network training: 
\begin{equation}\label{eq:lsimple_supp}
    \mathcal{L}_{\text{simple}} = E_{x_0 \sim q(x_0), t \sim U[1,T]}||\epsilon - \epsilon_{\theta}(\x_t, t) ||^2_2.
\end{equation}
In Eq.~\ref{eq:x_t_supp}, ${\alpha_t = 1 - \beta_t}$, ${\bar{\alpha_t} = \prod_{i=1}^T\alpha_i}$, $\beta_t$ define the variance schedule for $t \in\{1, \dots T\}$, and $\epsilon \sim \mathcal{N}(0, \mathbf{I})$.

We found that optimizing $\epsilon_\theta$ to approximate the noise $ \epsilon$ (Eq.~\ref{eq:lsimple_supp}) is equivalent to directly minimizing the local rotation error.
\begin{lemma} \label{eq:lemma1_sup}
Let $\mathcal{L}(x, x')=||x - x'||^2$ be the local rotation error loss between motion sequences $x$ and $x'$, where $x'$ is an estimate of $x$. Then, optimizing the $\mathcal{L}_{\text{simple}}$ loss is equivalent to optimizing $\mathcal{L}$. 
\end{lemma}

\begin{proof}
Let the rotation loss be
\begin{equation}\label{eq:rot}
    \mathcal{L}(x, x')=||x - x'||^2.
\end{equation}
Considering that $x_t$ for any single step in the DDPM is generated with Eq.~\ref{eq:x_t_supp}, we can solve for $x$ from this equation. 
Similarly, since the DDPM model generates an estimate of $\epsilon$, we can generate the estimate $x'$ by replacing $\epsilon$ with $\epsilon_\theta$. 
Hence,
\begin{equation}\label{eq:estimates}
\begin{split}
    x &= \frac{1}{\sqrt{\bar{\alpha}_t}} (\x_t - \sqrt{1 - \bar{\alpha}_t}\epsilon), \\
    x' &= \frac{1}{\sqrt{\bar{\alpha}_t}} (\x_t - \sqrt{1 - \bar{\alpha}_t}\epsilon_\theta(\x_t)).
\end{split}
\end{equation}

By combining Eq.~\ref{eq:estimates} in Eq.~\ref{eq:rot}, we compute
\begin{equation}
\begin{split}
    \mathcal{L}(x, x') &= ||x - x'||^2 \\
    &= \bigg|\bigg| \frac{1}{\sqrt{\bar{\alpha}_t}} (\x_t - \sqrt{1 - \bar{\alpha}_t}\epsilon) \\
    &- \frac{1}{\sqrt{\bar{\alpha}_t}} (\x_t - \sqrt{1 - \bar{\alpha}_t}\epsilon_\theta(\x_t)) \bigg|\bigg|^2 \\
    &= \bigg|\bigg| \sqrt{\frac{1 - \bar{\alpha}_t}{\bar{\alpha}_t}} (\epsilon - \epsilon_\theta(\x_t)) \bigg|\bigg|^2 \\
    &= \frac{1 - \bar{\alpha}_t}{\bar{\alpha}_t} \big|\big| \epsilon - \epsilon_\theta(\x_t) \big|\big|^2 
\end{split}
\end{equation}

Therefore,
\begin{equation}
    \mathcal{L}(x, x') = \frac{1 - \bar{\alpha}_t}{\bar{\alpha}_t} \mathcal{L}_{\text{simple}}(\epsilon, \epsilon_\theta(\x_t)),
\end{equation}
showing that minimizing the local rotation and the simple loss is equivalent to a scaling factor.
\end{proof}

\twocolumn[{%
\renewcommand\twocolumn[1][]{#1}%
\maketitle
\begin{center}
    \centering
    \includegraphics[width=0.9\linewidth]{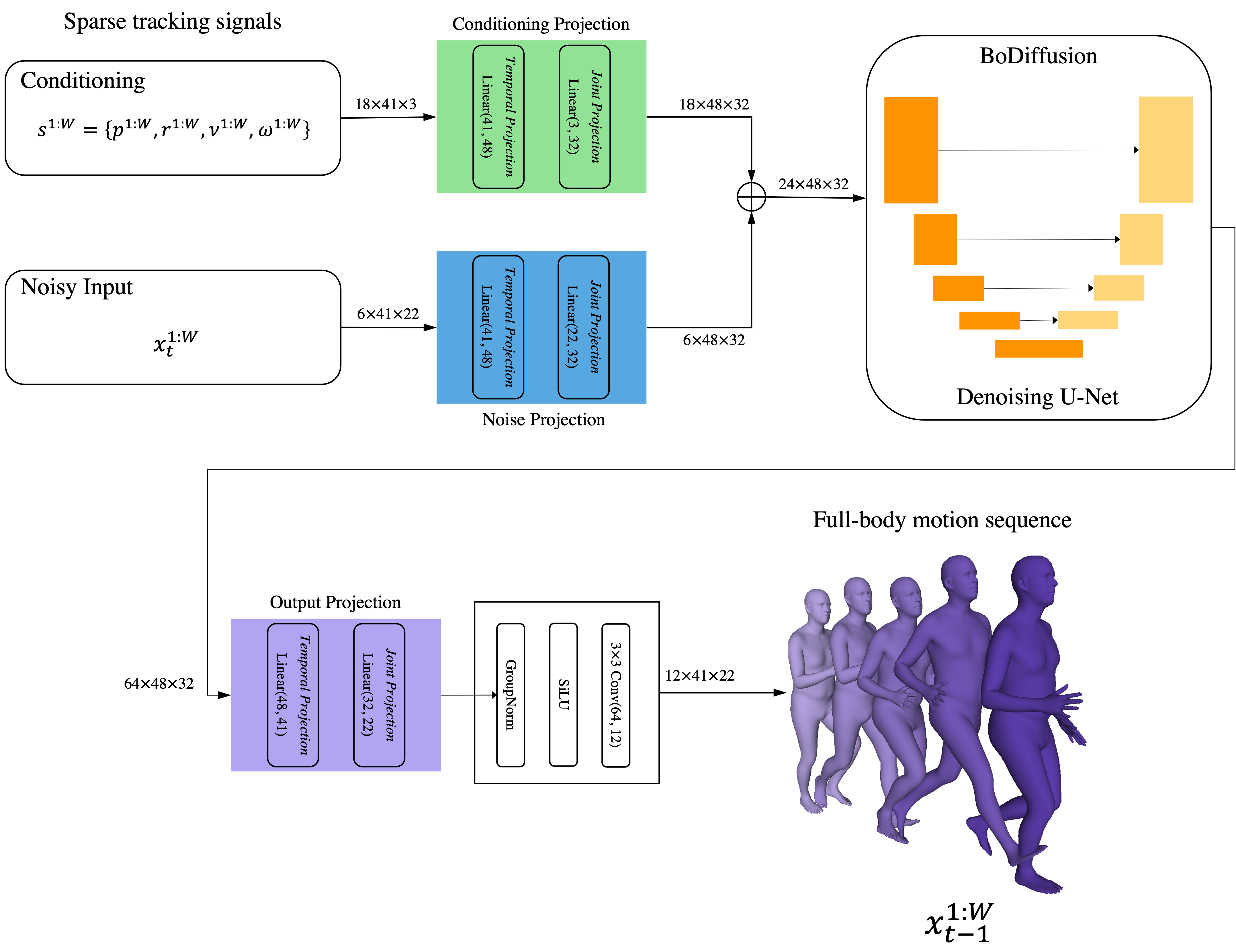}
    \captionof{figure}{
    \textbf{Overview of~\MODELNAME-UNet.} Here we show a version of \MODELNAME\, based on the U-Net architecture.
    We condition the input signals after a conditioning projection at the top (green block). Similarly, we project the noisy input local rotations (blue block) to match the sizes from the conditioning pathway. After denoising by the U-Net, we perform a final projection to the original space of full body motions (purple block). The output estimates $\epsilon_{\theta}$ and $\Sigma_{\theta}$ that are used to compute the local rotations $x_{t-1}^{1:W}$ by sampling from $\mathcal{N}(x_{t-1}; \mu_{\theta}, \Sigma_\theta)$.  
    $\oplus$ is the operation of concatenation along the channels' dimension. The numbers next to the arrows denote the input and output dimensions for the corresponding blocks. 
    }
    \label{fig:initial_proj}
\end{center}
}]

\end{document}